\documentclass[10pt,twocolumn,letterpaper]{article}

\usepackage{iccv}      

\usepackage[dvipsnames]{xcolor}

\NewDocumentCommand{\todo}
{ mO{} }{\textcolor{magenta}{\textsuperscript{\textit{TODO}}\textsf{\textbf{\small[#1]}}}}

\definecolor{iccvblue}{rgb}{0.21,0.49,0.74}
\usepackage[pagebackref,breaklinks,colorlinks,allcolors=iccvblue]{hyperref}
\def\paperID{11955} 
\def\confName{ICCV}
\def\confYear{2025}

\title{FB-4D: Spatial-Temporal Coherent Dynamic 3D Content Generation with Feature Banks}

\author{
    \begin{tabular}{c}
        Jinwei Li\textsuperscript{*1,2},  
        Huan-ang Gao\textsuperscript{*1,2},  
        Wenyi Li\textsuperscript{1},  
        Haohan Chi\textsuperscript{1,2},  
        Chenyu Liu\textsuperscript{1,2},\\
        Chenxi Du\textsuperscript{2},  
        Yiqian Liu\textsuperscript{1,2}  
        Mingju Gao\textsuperscript{1},
        Guiyu Zhang\textsuperscript{1},
        Zongzheng Zhang\textsuperscript{1},\\ 
        Li Yi\textsuperscript{3},
        Yao Yao\textsuperscript{4}, 
        Jingwei Zhao\textsuperscript{5},
        Hongyang Li\textsuperscript{6},
        Yikai Wang\textsuperscript{7}, 
        Hao Zhao\textsuperscript{\textdagger 1} \vspace{0.4em}\\
        \textsuperscript{1}AIR, Tsinghua University \quad
        \textsuperscript{2}DCST, Tsinghua University\quad
        \textsuperscript{3}IIIS, Tsinghua University\\
        \textsuperscript{4}Nanjing University \quad
        \textsuperscript{5}Xiaomi Corporation \quad
        \textsuperscript{6}Shanghai AI Laboratory\\
        \textsuperscript{7}School of Artificial Intelligence, Beijing Normal University\vspace{0.4em}\\
        Project Page: \url{https://FB-4D.c7w.tech/}
    \end{tabular}
}

\begin{document}
\maketitle

\begingroup
\renewcommand\thefootnote{}
\makeatletter\def\Hy@Warning#1{}\makeatother
\footnotetext{
    \noindent
    \textsuperscript{1} {*}Indicates Equal Contribution
    \textsuperscript{\textdagger}Indicates Corresponding Author\\
    
}
\endgroup

\begin{abstract}
With the rapid advancements in diffusion models and 3D generation techniques, dynamic 3D content generation has become a crucial research area. However, achieving high-fidelity 4D (dynamic 3D) generation with strong spatial-temporal consistency remains a challenging task. Inspired by recent findings that pretrained diffusion features capture rich correspondences, we propose FB-4D, a novel 4D generation framework that integrates a \underline{F}eature \underline{B}ank mechanism to enhance both spatial and temporal consistency in generated frames. In FB-4D, we store features extracted from previous frames and fuse them into the process of generating subsequent frames, ensuring consistent characteristics across both time and multiple views.
To ensure a compact representation, the Feature Bank is updated by a proposed dynamic merging mechanism. Leveraging this Feature Bank, we demonstrate for the first time that generating additional reference sequences through multiple autoregressive iterations can continuously improve generation performance. 
Experimental results show that FB-4D significantly outperforms existing methods in terms of rendering quality, spatial-temporal consistency, and robustness. It surpasses all multi-view generation tuning-free approaches by a large margin and achieves performance on par with training-based methods. Our code and data are released \href{https://github.com/dufengfeng/FB-4D}{here}.
\end{abstract}
\vspace{-15pt}    

\section{Introduction}
Dynamic 3D content generation, commonly known as 4D generation, involves not only creating the 3D geometry and appearance of objects but also capturing their motion over time within 3D space. Recently, interest in 4D content generation has surged \cite{jiang2023consistent4d, zeng2025stag4d, xie2024sv4d, singer2023text, jiang2025animate3d, chu2025dreamscene4d, yu20244real, bahmani2024tc4d,li20244k4dgen,zhao2024genxd,sun2024eg4d}, driven by its significance in applications such as autonomous vehicle simulation, gaming and film production, digital avatar creation, and immersive video experiences.

\begin{figure}
    \centering
    \includegraphics[width=0.75\linewidth]{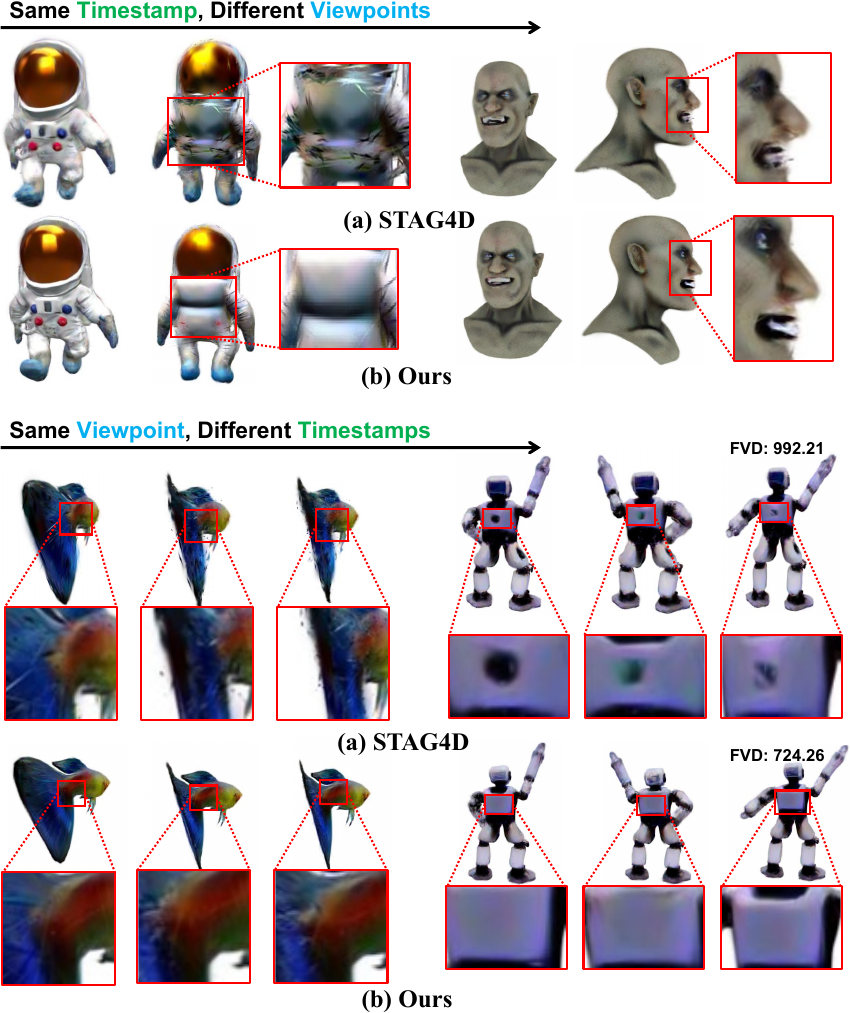}
    \caption{Compared to STAG4D \cite{zeng2025stag4d}, our method enhances spatial consistency (top row, better texture alignment across different viewpoints) and temporal consistency (bottom row, with fewer floaters and smoother motion transitions) by leveraging implicit correspondences in diffusion features. This translates to a superior FVD score than STAG4D (724.26 v.s. 992.21).}
    \label{fig:gd1}
    \vspace{-15pt}
\end{figure}

The primary challenge in dynamic 3D generation lies in jointly modeling spatial-temporal correspondences. 
To tackle this, borrowing from static 3D \cite{liu2023zero,shi2023zero123++,long2024wonder3d,li2024era3d}, current approaches typically employ a two-stage process. The first stage involves generating multiple anchor views of sequences, while the second stage uses these anchor views to train a 4D representation with score distillation \cite{liu2023syncdreamer}.
For the first stage, there are two primary approaches.
\textbf{(I) Training-based methods}, such as SV4D \cite{xie2024sv4d} and 4Diffusion \cite{zhang20244diffusion}, solve this task in a data-driven manner by modeling the joint latent space of views and frames, and conducting view and frame attention interleaved.
However, the length of videos and the number of views that can be processed with these methods are constrained by GPU memory limitations, such as the need for 80GB GPUs, which are not accessible to content creators using typical workstations.
\textbf{(II) Training-free methods} disentangles the problem by relying on priors from pre-trained video and multi-view generative models \cite{shi2023mvdream}.
A notable example is STAG4D \cite{zeng2025stag4d}, which takes a single image as input, first generating a video using a video diffusion model \cite{blattmann2023stable} and then using a multi-view diffusion model \cite{shi2023zero123++} to produce multi-view images conditioned on each video frame. 
However, as shown in Fig.~\ref{fig:gd1}(a), generated frames of these approaches often suffer from inconsistencies in both the spatial dimension (e.g., textures misaligned with frontal views) and the temporal dimension (e.g., floaters disrupting motion continuity), as they model object motion and novel view synthesis separately.



\textbf{No Joint Modeling. Re-use Past Features for the Present.}
Recent studies have shown that diffusion features \cite{tang2023emergent, luo2024diffusion}, captured during denoising, contain rich correspondences across images. Building on this insight, we propose FB-4D, a novel 4D generation framework that \textit{employs a spatial-temporal feature bank} to address the correspondence learning challenge by leveraging pretrained representations from diffusion-based multi-view generation models.
To ensure the feature bank functions effectively, simply storing intermediate features from past frames or using a sliding window approach yields suboptimal results. Instead, we propose a dynamic mechanism for managing the feature bank’s contents by continuously merging redundant information from new and stored features. This approach enables the preservation of the most representative features while maintaining a consistent bank size over time.

\textbf{The more you see in 2D, the more you perceive in 3D.}
Enhancing 3D generation by introducing additional 2D anchors has proven effective \cite{han2024more}. A straightforward approach to this is through autoregressive generation, which produces additional reference images from a generated view to optimize the neural field. However, for 4D generation, this method alone is problematic, as inconsistencies introduced at each generation step can accumulate across iterations. With the integration of our proposed feature bank mechanism, however, we demonstrate for the first time that \textit{generating additional reference sequences through multiple autoregressive iterations} can reliably enhance downstream performance. And this approach can be further strengthened by progressively selecting reference views.

FB-4D requires \textit{neither training nor fine-tuning} of any multi-view or video generation model, yet it effectively addresses the challenge of correspondence matching across multiple frames and views, as illustrated in Fig.~\ref{fig:gd1}(b). Through extensive qualitative and quantitative experiments, we show that FB-4D achieves state-of-the-art results on the well-established dynamic 3D content generation benchmark, Consistent4D \cite{jiang2023consistent4d}, outperforming all training-free methods by a substantial margin and matching the performance of the training-based SV4D \cite{xie2024sv4d}. Furthermore, we conduct in-depth experiments to explore how, why, and where the proposed feature bank mechanism is most effective. In summary, our contributions are as follows:
\begin{itemize}
    \item We propose FB-4D, a novel framework that leverages the feature bank mechanism to enable training-free, efficient correspondence modeling across multiple frames and views for dynamic 3D content generation tasks.
    \item We conduct extensive experiments and analyses to provide in-depth insights into the feature bank mechanism, detailing how, why, and where it operates effectively.
    \item Our method achieves state-of-the-art results on the 4D generation benchmark, significantly outperforming all training-free methods and matching the performance of training-based approaches like SV4D \cite{xie2024sv4d}.
\end{itemize}

\section{Related Works}

\textbf{3D Generation and Video Generation.}
The domain of static 3D generation has drawed significant attention due to the advancements in 3D representation learning methods \cite{kerbl20233dgs, muller2022instant, xu2022point, xie2024physgaussian, guedon2024sugar, yu2024mip, yang2024diffusion, liu2024rip}. 
Despite that various research endeavors aim to enhance score distillation loss \cite{yi2023gaussiandreamer, tang2023dreamgaussian, shi2023mvdream, wang2024prolificdreamer, li2023sweetdreamer, weng2023consistent123, chen2024text, sun2023dreamcraft3d, sargent2023zeronvs, liang2024luciddreamer, zhou2024dreampropeller, guo2023stabledreamer} and facilitate generation in a feed-forward manner \cite{hong2023lrm, jiang2023leap, wang2023pf, zou2024triplane, wei2024meshlrm, tochilkin2024triposr}, our focus is on approaches that generate dense multi-view images with sufficient 3D consistency, subsequently reconstructing 3D content from these images \cite{liu2023zero, liu2023syncdreamer, long2024wonder3d, voleti2025sv3d, ye2024consistent, karnewar2023holofusion, li2023instant3d, shi2023toss, shi2023zero123++, wang2023imagedream, liu2024one_a, liu2024one_b}. Our research adheres to this paradigm, but extends it by generating consistent multi-view videos (instead of images) and subsequently reconstructing the 4D object.
For video generation, thanks to the powerful generative modeling ability of diffusion \cite{blattmann2023stable,rombach2022high,zhang2023adding,gao2024scp,li2024fairdiff,zhang2024ctrl,ni2025straight,chen2024ultraman,xu2024diffusion,li2025avd2}, video diffusion models have shown exceptional performance with consistent geometry and realistic motions \cite{ho2022video, voleti2022mcvd, blattmann2023align, blattmann2023stable, he2022latent, singer2022make, guo2023animatediff, bahmani2024vd3d, he2024cameractrl, wang2024motionctrl}. Their robust generalization stemming from training on extensive image and video datasets prompts us to leverage them for text-to-video or image-to-video generation, creating the initial reference sequence.

\textbf{4D Generation.}
The generation of 4D content, or dynamic 3D content, is enabled by the advancements in 4D representation learning \cite{blattmann2023stable, wang2023modelscope}.
Given the challenges of applying score distillation sampling across multiple frames to directly distill motion knowledge from video diffusion models \cite{xie2024sv4d}, existing text-to-4D \cite{singer2023text, ling2024align, bahmani20244d} and image-to-4D methods \cite{zhao2023animate124} often exhibit suboptimal appearance quality. 
Consequently, many approaches have shifted towards video-to-4D generation \cite{jiang2023consistent4d, ren2023dreamgaussian4d, yi2023gaussiandreamer}. 
These methods either models joint latent of view and time \cite{xie2024sv4d, zhang20244diffusion} or utilize contemporary multiview diffusion models \cite{liu2023zero, liu2023syncdreamer, shi2023zero123++} to compute the SDS loss \cite{ren2023dreamgaussian4d, jiang2023consistent4d, wu2025sc4d} on generated per-frame multi-view images \cite{zeng2025stag4d, yang2024diffusion} as a supervisory signal \cite{jiang2023consistent4d, zeng2025stag4d}. 

\section{Preliminaries and Our Motivation}

\label{pre_knowledge}
\textbf{Overview of STAG4D.} 
As discussed in the introduction, we abandon jointly modeling the latents of frames and views, opting instead to use past features for the generation of the current frame. To understand this approach, we first delve into STAG4D, a framework designed for high-fidelity 4D generation by integrating pre-trained diffusion models with dynamic 3D Gaussian splatting. The framework operates in two stages. Given \textbf{Single-view video frames} are used as input, \textbf{stage 1} employs a multi-view diffusion model, such as Zero123++ \cite{shi2023zero123++}, to generate multi-view frame sequences from the reference frame sequence.  

For convenience, we define the input viewpoint as $v=0$ and the output viewpoints as $v=1:6$. Let the input frame sequence contain $T$ frames, denoted as $F^{[v=0,t=1:T]}$. After passing through the multi-view diffusion model, the output frame sequence can be represented as $F^{[v=1:6,t=1:T]}$. The entire multi-view diffusion process can be formulated as:
\begin{equation}
    F^{[v=1:6,t=1:T]} = \text{Generator} \left( F^{[v=0,t=1:T]} \right),
\end{equation}
where the Generator denotes the multi-view diffusion process.
\textbf{Stage 2} utilizes score distillation sampling to refine and optimize the 4D Gaussians \cite{zeng2025stag4d}, given the reference frame sequence and multi-view generated sequences as input. For each rendering viewpoint, the closest viewpoint from the multi-view generated sequences is determined based on proximity, denoted as $i$, where \( i \) ranges from 1 to 6. The multi-view score distillation loss is formulated as:
\begin{equation}
    L_{\text{MVSDS}} = \lambda_1 L^{i}_{\text{SDS}} + \lambda_2 L^0_{\text{SDS}},
\end{equation}
where $\lambda_1$ and $\lambda_2$ are weighting factors.

\begin{figure}[t]
    \centering
    \includegraphics[width=1\linewidth]{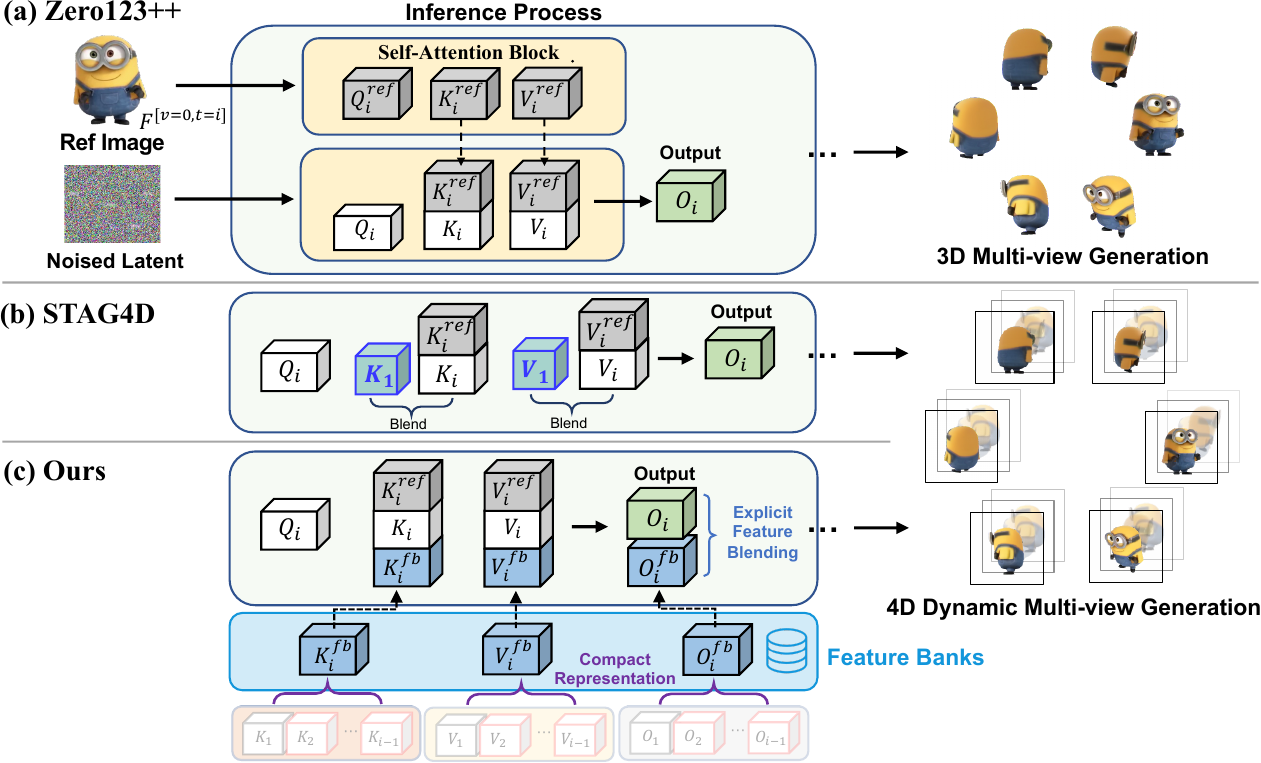}
    \caption{\textbf{Paradigm comparison of our method with previous works.} (a) Zero123++ uses a \textbf{\textcolor[HTML]{ffcc14}{dual-branch self-attention mechanism}}, one branch for the reference image and the other for noised latent, with the reference image enhancing spatial consistency. (b) STAG4D adds \textbf{\textcolor{blue}{key-value information from the first frame}} to improve frame consistency. (c) Our method introduces a feature bank that stores a \textbf{\textcolor[HTML]{9c43ea}{compact representation}} deduced from the previous \( i-1 \) frames, with \textbf{the same size} as a \textit{single} frame but contains richer feature temporally. This bank is used to generate \( F^{[v=0,t=i]} \). After obtaining the output \( \textbf{O}_\textbf{i} \) from the self-attention block, we blend it with the stored \( \textbf{O}_\textbf{i}^{\text{fb}} \) from the feature bank to enhance temporal consistency (detailed in Sec.~\ref{sec32}).}


    \label{fig:preliminary}
\end{figure}

\begin{figure*}
    \centering
    \includegraphics[width=0.9\textwidth]{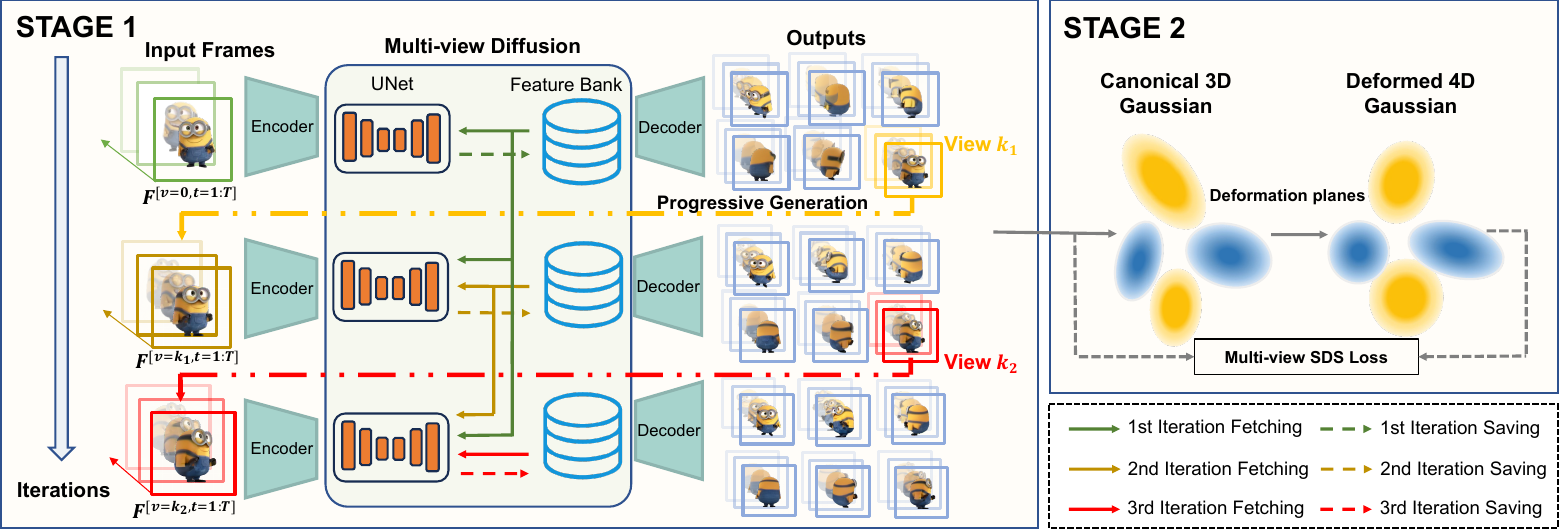}
    \caption{\textbf{Overall pipeline of video-to-4D generation in FB-4D.} Given a single-view video input, we integrate a multi-view generative diffusion model \cite{shi2023zero123++} with a feature bank to enhance spatial and temporal consistency, implicitly modeling the correspondence across views and timestamps (detailed in Sec.~\ref{sec32}). We perform multiple iterations of generation, where each generated view is used as input for the next iteration. For example, in the first iteration, we use \( F^{[v=0,t=1:T]} \) to generate multi-view sequences \( F^{[v=1:6,t=1:T]} \) and in the second iteration, we select the input view progressively (detailed in Sec.~\ref{sec33}). The input for this iteration becomes \( F^{[v=k_1,t=1:T]} \), where \( k_1 \in [1,6] \). This iterative process enables interaction across feature banks, refining the multi-view sequences over time. The generated sequences maintain high consistency in both spatial and temporal dimensions. After several iterations, we finally train a deformable 3D Gaussian to represent the 4D model.}
    \label{fig:gd2}
    \vspace{-10pt}
\end{figure*}

In Stage 1, STAG4D builds upon Zero123++, which incorporates the reference image into U-Net self-attention blocks through a reference branch and a noised latent branch. Given an input \( F^{[v=0,t=i]} \), the self-attention keys, queries, and values from the reference branch are denoted as \( \{{\textbf{Q}_\textbf{i}}^{\text{ref}}, {\textbf{K}_\textbf{i}}^{\text{ref}}, {\textbf{V}_\textbf{i}}^{\text{ref}}\} \), while those from the noised latent branch are denoted as \( \{{\textbf{Q}_\textbf{i}}, {\textbf{K}_\textbf{i}}, {\textbf{V}_\textbf{i}}\} \), producing an output denoted as \( {\textbf{O}_\textbf{i}} \) (see Fig.~\ref{fig:preliminary} (a)). While this design enhances spatial consistency, it processes each frame independently, leading to temporal inconsistencies. To address this, STAG4D extends keys and values in self-attention layers with those of the first input frame \( F^{[v=0,t=1]} \), ensuring greater temporal coherence (see Fig.~\ref{fig:preliminary} (b)).

However, as shown in Fig.~\ref{fig:gd1} (b), temporal inconsistency issues remain. We believe that considering only the first input frame \( F^{[v=1:6,t=1]} \) as a condition during the generation of the subsequent \( i \)-th frame \( F^{[v=1:6,t=i]} \) is insufficient for achieving better temporal consistency. (Take the rear view $v=3,4$ for example. Only conditioned on the first frame cannot ensure a smooth transition through time.)
Instead, the conditions from all previous time, \( F^{[v=1:6,t=1:(i-1)]} \), should be incorporated during the generation of \( F^{[v=1:6,t=i]} \). Therefore, we extended the design of the attention blocks to incorporate all the relevant features from previously generated frames (see Fig.~\ref{fig:preliminary}(c)).



\section{Method}
\subsection{Overview}

The overall pipeline, as shown in Fig.~\ref{fig:gd2}, illustrates the process of video-to-4D generation using FB-4D method. 
To address the spatial-temporal inconsistencies, we integrate the multi-view generation model \cite{shi2023zero123++} with a feature bank, implicitly modeling correspondences across views and timestamps.
In Sec.~\ref{sec32}, we explore this design by explaining how the feature bank is updated and fetched, detailing the compact representation and feature blending in Fig.~\ref{fig:preliminary}.
In Sec.~\ref{sec33}, we introduce a novel approach that combines auto-regressive generation with our feature bank to generate more multi-view image sequences. For iteration index \( j \), one of the output sequences from the previous iterations, \( F^{[v=1:(6j-6), t=1:T]} \), serves as the input for the next iteration. Specifically, \( F^{[v=k, t=1:T]} \) is selected, where \( k \) belongs to the range \( [1, 6j-6] \). Furthermore, we propose a new progressive viewpoint selection strategy to enhance the generation process.
After generating multi-view, multi-frame images, these are used to train a deformable 3D Gaussian field, enabling novel view rendering at continuous spatial positions and timestamps.


\subsection{Implicit Correspondence Modeling with Feature Banks}
\label{sec32}
\textbf{Hooking Self-Attention Layers with Feature Bank: Overview.} 
To store features from past frames, we propose integrating a newly designed feature bank module in the self-attention layers of multi-view generation models.
For clarity, when processing the \( i \)-th input frame \( F^{[v=0, t=i]} \), we denote the current feature bank set as \( \textbf{S}_\textbf{i}^{\text{fb}} \), where \( \textbf{S} \in \{\textbf{K}, \textbf{V}, \textbf{O}\} \). The shape of an element in the set, e.g., \( \mathbf{\textbf{K}}_{\textbf{i}} \), is \( \frac{hw}{s^2} \times D \), where \( \frac{hw}{s^2} \) is the token count and \( D \) is the feature dimension. During the denoising process, the feature bank utilizes the feature set \( \textbf{S}_\textbf{i} \) from the noised latent branch, i.e., \( ({\textbf{K}_\textbf{i}}, {\textbf{V}_\textbf{i}}, {\textbf{O}_\textbf{i}}) \), to update itself (Note that the notation \( t \) represents the index of the frame being generated. To enhance adaptability and performance, we employ distinct feature banks for different inference steps in the diffusion process and omit the notation for it).





The feature bank mechanism is integrated into all self-attention blocks, and we evaluate the impact of this choice in our experiments (see Table \ref{tab:layers}). Next, we provide a detailed explanation of the processes for fetching and updating the feature bank, during which we implemented compact representation and feature blending. (see Fig.~\ref{fig:preliminary} (c))

\textbf{Updating the Feature Bank: A Compact Representation.} A straightforward update method operates in a queue-like fashion, where the maximum number of past frames is fixed. When the feature bank reaches its capacity, new frame features are added, and the oldest are removed. While maintaining a fixed size, this limits current frame's access to earlier features. Increasing the window size to include more frames requires more memory and may introduce redundancy, diminishing its effectiveness (see Table \ref{tab:queue}).

\begin{figure}
    \centering
    \includegraphics[width=1\linewidth]{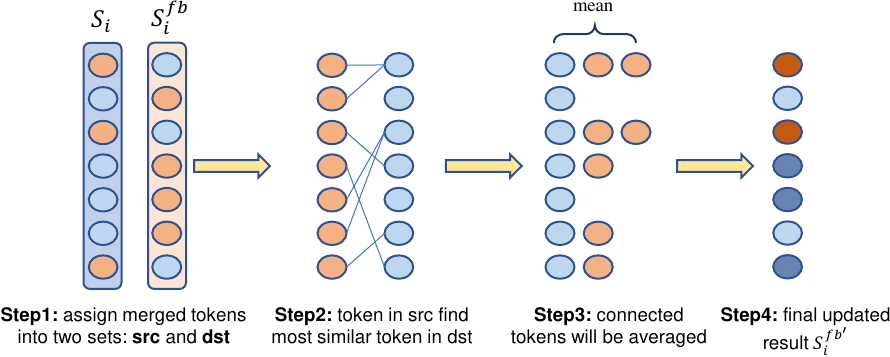}
    \caption{\textbf{Illustration of the compact feature fusion in the updating process}. The tokens in the feature bank are continuously updated through the aforementioned four steps, effectively integrating past features while ensuring that the size of the feature bank remains constant.}
    \label{fig:gd3}
\end{figure}

To address this, we propose a dynamic update strategy for the feature bank that retains essential features from past frames while efficiently controlling its size. 
We employ an efficient greedy strategy to merge the features in \( \textbf{S}_\textbf{i} \) and \( \textbf{S}_\textbf{i}^{\text{fb}} \) \cite{yang2024model, li2024training, biggs2024diffusion, bolya2022token}. Specifically, the feature bank update procedure is outlined as follows: (illustrated in Fig.~\ref{fig:gd3})

\begin{enumerate}
    \item We merge the current frame's \(\mathbf{\textbf{S}}_\textbf{i}\) with \(\mathbf{\textbf{S}}^{\text{fb}}_\textbf{i}\) from the feature bank in the token dimension to form \(\mathbf{\textbf{S}}^{\text{merge}}_\textbf{i}\), and then randomly split it into two subsets, \(\text{src}\) and \(\text{dst}\), to maintain diversity without clustering's high cost.
    \item For each feature \(\text{src}_p\) in \(\text{src}\), we find the most similar feature \(\text{dst}_q\) in \(\text{dst}\) using cosine similarity.
    \item We then fuse features by averaging those matched to \(\text{dst}_q\), resulting in the updated feature set \( \mathbf{\textbf{S}}_{\textbf{i}}^{\text{fb}'} \).

\end{enumerate}
Through this approach, we effectively control the size of the feature bank while integrating all past features. As a result, \(\mathbf{\textbf{S}}_{\textbf{i}}^{\text{fb}}\) serves as a compact representation of the previous \(i-1\) frames \( F^{[v=0, t=1:(i-1)]} \).


For initialization of the feature bank, we utilize a warm-up procedure before performing the inference. Specifically, we use the first five frames to initialize the feature bank. 
This strategy enables the feature bank to accumulate a more diverse and representative set of features.

\begin{figure}
    \centering
    \includegraphics[width=0.8\linewidth]{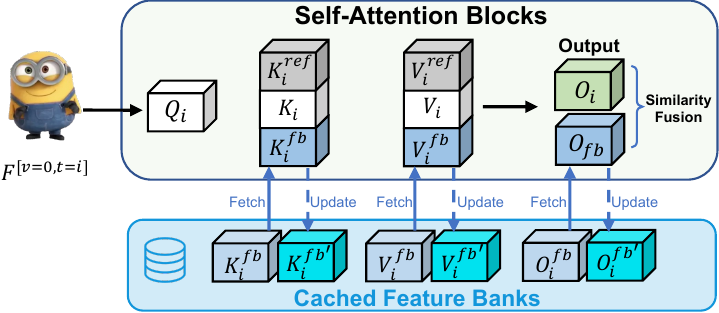}
    \caption{\textbf{Illustration of feature bank fetching and updating.} In the multi-view inference process of frame \( F^{[v=0, t=i]} \), we first retrieve the corresponding feature set for computation and then utilize the keys, values, and outputs (\(\mathbf{\textbf{K}}_\textbf{i}\), \(\mathbf{\textbf{V}}_\textbf{i}\), \(\mathbf{\textbf{O}}_\textbf{i}\)) from the self-attention blocks to update the feature bank.}
    \label{fig:fb-final}
\end{figure}

\textbf{Fetching From the Feature Bank.} \textit{(I) Fetching Key and Value.}  
During self-attention computation for frame \( F^{[v=0, t=i]} \), the model retrieves the key \(\mathbf{K}_{\text{fb}}\) and value \(\mathbf{V}_{\text{fb}}\) from the feature set \(\mathbf{\textbf{S}}_{\textbf{i}}^{\text{fb}}\), efficiently conditioning on past frames.  
To utilize these features, we extend reference attention \cite{shi2023zero123++} (Fig.~\ref{fig:preliminary}), formulated as follows (see Fig.~\ref{fig:fb-final}):
\begin{equation}
\resizebox{0.9\linewidth}{!}{
\begin{math}
\begin{gathered}
\text{SelfAttn}(\mathbf F^{[v=0, t=i]}) = \text{softmax}\left(\frac{\mathbf {\textbf{Q}_\textbf{i}} [\mathbf {\textbf{K}_\textbf{i}}, \mathbf K_{\text{ref}}, \mathbf K_{\text{fb}}]^T}{\sqrt{d}}\right) [\mathbf {\textbf{V}_\textbf{i}}, \mathbf V_{\text{ref}}, \mathbf V_{\text{fb}}],
\label{eq:anti-se}
\end{gathered}
\end{math}
}
\end{equation}
\begin{equation}
\mathbf {\textbf{O}_\textbf{i}} = \text{MLP}_\text{to\_out}(\text{SelfAttn}(\mathbf F^{[v=0, t=i]})),
\end{equation}


\textit{(II) Fetching Output.} 
In addition to the key and value, the output $\mathbf{O}_{\text{fb}}$ is also used in a similarity fusion manner for the current generation (see feature blending in Fig.~\ref{fig:preliminary} (c)). For each token \(\alpha\) in \(\mathbf{O}_i\), we use cosine similarity to find the most similar token \(\beta\) in \(\mathbf{O}_{\text{fb}}\). The final output element \(\mathbf{O}'_i(\alpha)\) can then be expressed as,
\begin{equation}
\mathbf{O}'_i(\alpha) = (1 - \lambda) \cdot \mathbf{O}_i(\alpha) + \lambda \cdot \mathbf{O}_{\text{fb}}(\beta),
\end{equation}
where \(\lambda\) is a hyperparameter used to control the strength of the fusion. 
We validate the necessity of fetching key-values and outputs separately in Tab.~\ref{tab:featurebank_validation}. Additionally, to prevent excessive fusion with past frames, a threshold \(\tau\) is set to generate a mask, ensuring feature fusion occurs only when the cosine similarity is high. This preserves important details and prevents the current frame from becoming overly similar to past frames, avoiding reduced motion magnitude.

\subsection{Progressive Generation in Multiple Iterations}
\label{sec33}





We compared three multi-iteration methods, distinguished by colors: \textbf{\textcolor[HTML]{997500}{brown}},  \textbf{\textcolor[HTML]{14b32e}{green}} and \textbf{\textcolor[HTML]{b034e0}{purple}}.

\begin{table}[ht]
\centering
\resizebox{\linewidth}{!}{ 
\footnotesize
\begin{tabular}{c c c c c c}
\toprule
\textbf{Method} & \textbf{viewpoint} & \textbf{F-B} & \textbf{Iter 1} & \textbf{Iter 2} & \textbf{Iter 3}\\
\midrule
\textbf{\textcolor[HTML]{997500}{(a)}} & random & \(\times\) & 881.17 & 902.28 \textsuperscript{\textcolor[HTML]{997500}{+21.11}}& 930.06 \textsuperscript{\textcolor[HTML]{997500}{+48.89}}\\
\textbf{\textcolor[HTML]{14b32e}{(b)}} & random & \checkmark & 784.36  & 774.71 \textsuperscript{\textcolor[HTML]{14b32e}{-9.65}}& 772.05 \textsuperscript{\textcolor[HTML]{14b32e}{-12.31}}\\
\textbf{\textcolor[HTML]{b034e0}{(c)}} & progressive & \checkmark & 784.36  & 756.90 \textsuperscript{\textcolor[HTML]{b034e0}{-27.46}}& 728.86 \textsuperscript{\textcolor[HTML]{b034e0}{-55.50}}\\
\bottomrule
\end{tabular}
}
\caption{\textbf{FVD comparison in multi-iteration settings.} F-B indicates feature bank usage. The \textbf{\textcolor[HTML]{997500}{brown line (a)}} represents random selection of previous outputs. The \textbf{\textcolor[HTML]{14b32e}{green line (b)}} improves quality with feature banks. The \textbf{\textcolor[HTML]{b034e0}{purple line (c)}} further enhances stability and clarity through progressive generation.}
\label{tab:contrast_3ways}
\end{table}

\textcolor[HTML]{997500}{\textbf{The More You See in 2D, the More You Perceive in 3D?}}  
SAP3D \cite{han2024more} showed that increasing input images from diverse viewpoints during deformable 3D Gaussian training enhances generation quality. Naturally, we formulate an auto-regressive generation process, using generated viewpoints as new inputs for Zero123++ \cite{shi2023zero123++}, and generate more anchor sequences. 
However, this yielded suboptimal results.  
More iterations and additional inputs greatly increase FVD in 4D generation for the baselines (Tab.~\ref{tab:contrast_3ways}\textbf{\textcolor[HTML]{997500}{(a)}}).

We argue that the issue persists in maintaining multi-view and frame-to-frame consistency. Specifically, as the number of iterations increases, spatial consistency tends to degrade. This accumulated error of inconsistencies across iterations introduces more confusion in the training process of the 3D deformable Gaussian model.

\textcolor[HTML]{14b32e}{\textbf{Utilizing Our Feature Banks in Multiple Iterations.}}
For each iteration, due to the different input viewpoints, we utilize distinct feature banks to focus on more diverse features. For clarity, we define the feature set in the feature bank of the $j$-th iteration as $\mathbf{\textbf{S}}_{\text{fb}}^j$, where, for example, $\mathbf{K}_{\text{fb}}$ becomes $\mathbf{K}_{\text{fb}}^{j}$. 
To facilitate discussion, we define the input viewpoints for the second and third iterations as $k_1$ and $k_2$ respectively, where $k_1 \in [1,6]$ and $k_2 \in [1,12]$ and $k_1 \neq k_2$. For convenience, the first iteration input viewpoint, previously denoted as \(0\), is also represented as \(k_0\). For three iterations, the whole process is formulated as (where Generator denotes the multi-view diffusion process):
\begin{equation}
\begin{split}
F^{[v=1:6,t=1:T]} &= \text{Generator}(F^{[v=k_0,t=1:T]}), \\
F^{[v=7:12,t=1:T]} &= \text{Generator}(F^{[v=k_1,t=1:T]}), \\
F^{[v=13:18,t=1:T]} &= \text{Generator}(F^{[v=k_2,t=1:T]}),
\end{split}
\end{equation}

\begin{figure}[t]
    \centering
    \includegraphics[width=0.65\linewidth]{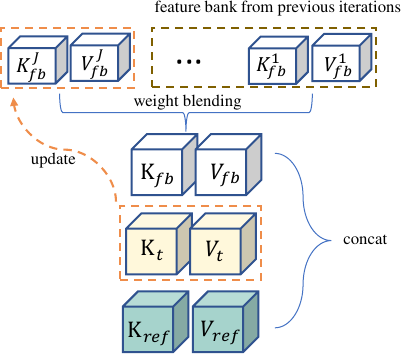}
    \caption{\textbf{Utilization of feature bank for generation in multiple iterations.} For fetching, keys and values from multiple feature banks are blended with weighted sum (Eq.\ref{eq:weighted_sum2}). For updating, newly predicted features are only updated into current feature banks.}
    \label{fig:gd4}
\end{figure}

\textit{Determinating the weights of feature banks in previous iteration.}
To address the accumulated inconsistency, we leverage the feature bank from previous iterations. By assigning corresponding weights to feature banks from different iterations, we perform key-value weight blending, as illustrated in Fig.~\ref{fig:gd4}, formulated as:
\begin{equation}
\label{eq:weighted_sum2}
    \mathbf{K}_{\text{fb}} = \sum_{j=1}^{J} w_j \cdot \mathbf{K}_{\text{fb}}^{j}, \quad 
    \mathbf{V}_{\text{fb}} = \sum_{j=1}^{J} w_j \cdot \mathbf{V}_{\text{fb}}^{j}.
\end{equation}
where \(J\) denotes the current iteration.
But how should these weights be determined? Intuitively, the greater the difference in viewpoints between the current iteration J and the previous iteration j, denoted as \( \Delta\theta(k_{J-1}, k_{j-1}) = |\text{azimuth}(k_{J-1}) - \text{azimuth}(k_{j-1})| \), the lower the weight assigned to the previous iteration.
Therefore, we define the weight for past feature banks as follows:  
\( w_j = \frac{\pi - \Delta\theta(k_{J-1}, k_{j-1})}{2\pi \cdot (J-1)} \).
And the weight assigned to the current feature bank is then given by \( w_J = 1 - \sum_{j=1}^{J-1} w_j \). 
In addition, in each iteration, we update only the feature bank of the current iteration to avoid excessive informational confusion across views also shown in Fig.~\ref{fig:gd4}.

As shown in Tab.~\ref{tab:contrast_3ways}\textbf{\textcolor[HTML]{14b32e}{(b)}}, during multiple iterations, leveraging the previous feature bank results in a significant reduction in FVD compared to not utilizing it. This demonstrates, for the first time, that generating additional reference sequences through multiple auto-regressive iterations can continuously improve downstream performance.

\begin{figure}[t]
    \centering

    \includegraphics[width=1\linewidth]{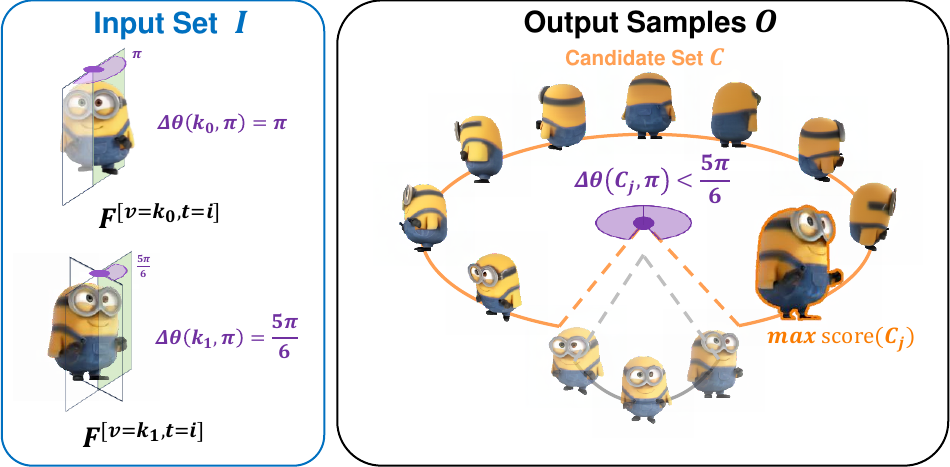}
    \caption{ \textbf{Visualization of our progressive iteration}. For the current iteration \( i=3 \), we compute the difference between the previous inputs \( F^{[v=k_0,t=i]} \) and \( F^{[v=k_1,t=i]} \) with \( \pi \), as shown in the left part. Among all the generated viewpoints shown in the right part, we select the one that is closer to the rear viewpoint (\( \pi \)) as the candidate set for the next iteration. And finally, we choose the highest-scoring viewpoint based on Eq.~\ref{eq8} as the final input.
}
    
    \label{fig:3d}
\end{figure}

Upon further investigation, we identified two issues with the multi-view image sequences generated after several iterations: (i) Randomly selecting viewpoints as new inputs may cause significant differences between the previous input viewpoints and the current ones, leading to a reduced utilization of knowledge in the feature bank; (ii) The quality of generated images from rear or side viewpoints is generally lower. To address these issues, we introduce a progressive generation mechanism.

\textcolor[HTML]{b034e0}{\textbf{Progressive Generation in Multiple Iterations.}} To address the aforementioned issues, we designed a progressive generation process. In this process, by the time of the \( J \)-th iteration, the difference between the input azimuth angle \( k_{J-1} \) and the rear viewpoint (\(\pi\) degrees) (denoted as \( \Delta\theta(k_{J-1}, \pi) \)) is gradually reduced over successive iterations, enabling smoother transitions between viewpoints and improving the consistency of generated images across different angles. This approach ensures that side and rear views are generated progressively, leading to higher consistency and better overall image quality.

To explain further, the input viewpoints from the previous \( J-1 \) iterations form a set \( I \), formulated as:
\[
I = \left\{ k_0, k_1, \dots, k_{(J-2)} \right\},
\]
representing all the input viewpoints before the current iteration. Here, \( k_{(J-2)} \) is the viewpoint closest to the rear (\(\pi\) degrees) in this set.
We also define the set of all viewpoints of output image sequences as \( O \), which contains \( v \) ranging from 1 to \( 6J-6 \). 
Next, we define a set of candidate viewpoints, \( C \), which consists of viewpoints from the remaining image sequences that are closer to the rear viewpoint than \( k_{(J-2)} \). The next input viewpoint for the current iteration is selected from \( C \).
This set can be expressed as:
\begin{equation}
C=\left\{ k \in O \setminus I \mid \Delta\theta(k, \pi) < \Delta\theta(k_{J-2}, \pi) \right\}
\end{equation}

\begin{figure*}[ht]     
    \centering     
    \includegraphics[width=0.8\linewidth ]{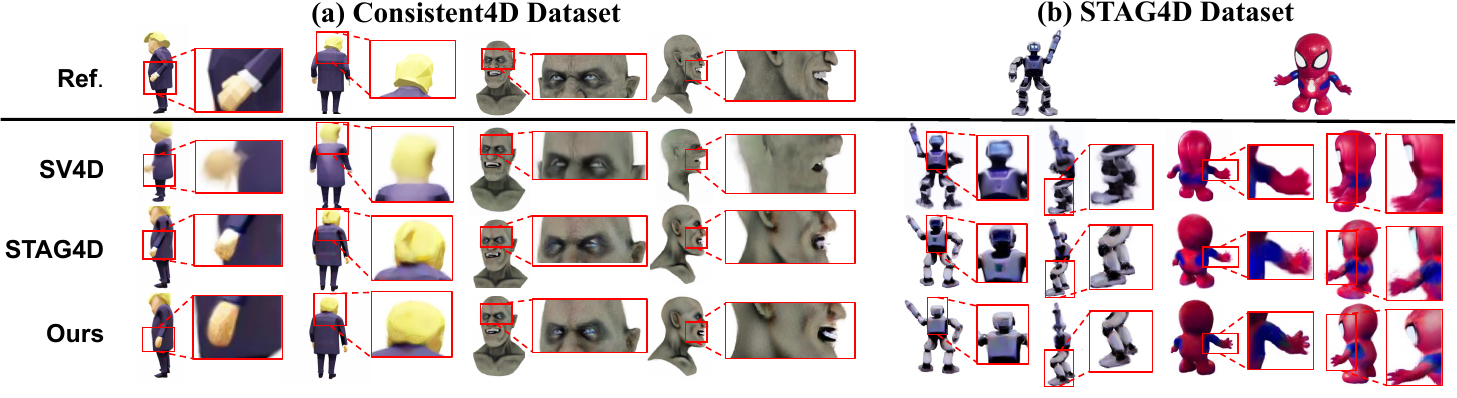}         
    \caption{\textbf{Comparison of 4D generation quality on (a) Consistent4D dataset and 
    (b) STAG4D dataset  across different methods.}}         
    \label{fig6}     
    \vspace{-3mm}   
\end{figure*}

\begin{figure*}[ht]     
    \centering     
    \includegraphics[width=0.7\linewidth ]{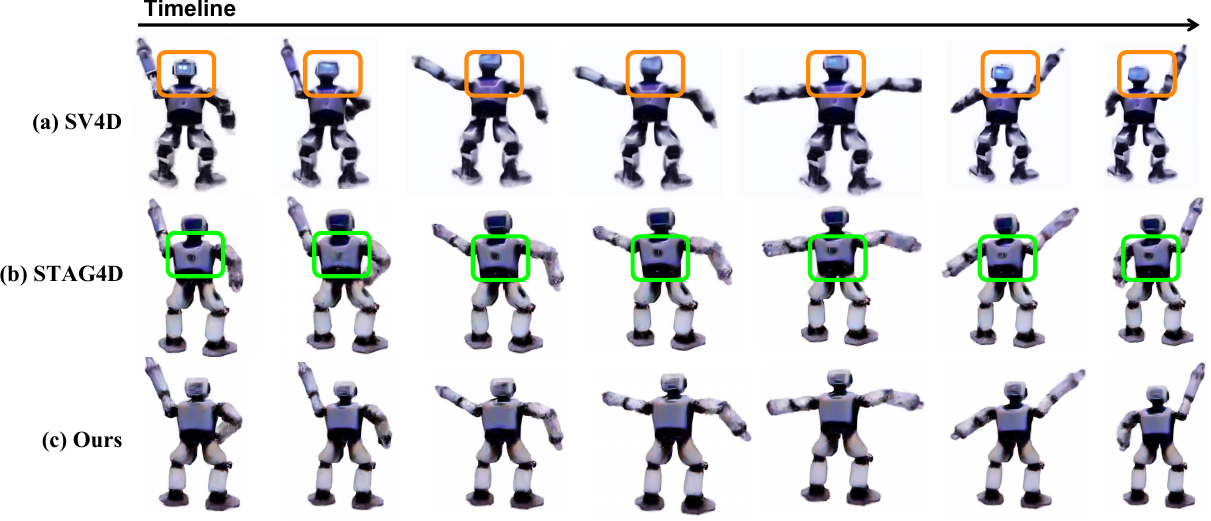}         
    \caption{\textbf{4D Generation Quality Comparison (Back View).} Both (a) SV4D and (b) STAG4D show rear view inconsistencies.}
    \label{movable}     
    \vspace{-3mm}   
\end{figure*}

After that, we calculate the similarity of each element in the candidate set $C_j \in C$ to the previous input viewpoints and select the viewpoint with the highest score for the next iteration input (as shown in Fig.~\ref{fig:3d}), written as,
\begin{equation}
\resizebox{0.9\linewidth}{!}{
    $\text{score}(C_j) = \sum_{p=0}^{J-2} \sum_{q=1}^{T} W_p \cdot \text{CLIP}(F^{[v=k_p,t=q]}, F^{[v=C_j,t=q]})$
}
\label{eq8}
\end{equation}

where \( W_p \) is the weight determined by the angular difference between the viewpoints, $W_p = (\pi - \Delta \theta(k_p, C_j)) / 2\pi$, with larger differences resulting in smaller weights.



The progressive generation approach maximizes feature bank usage for consistent multi-view image sequences and produces more high-quality sequences from different viewpoints. With this technique introduced, we are able to increase both the quantity and diversity of the generated images while explicitly ensuring their quality. As shown in the Tab.~\ref{tab:contrast_3ways}\textbf{\textcolor[HTML]{b034e0}{(c)}}, our progressively generated method has achieved favorable results against \textbf{\textcolor[HTML]{997500}{(a)}} and \textbf{\textcolor[HTML]{14b32e}{(b)}}.


\section{Experiments}
\subsection{Setup}


\textbf{Dataset and Metrics.} We conduct experiments using the Consistent4D dataset \cite{ye2024consistent}, which includes multi-view videos of seven dynamic objects. We evaluate using three metrics: CLIP, LPIPS, and FVD \cite{ye2024consistent}. CLIP and LPIPS assess image-level semantic and perceptual similarity, while FVD evaluates both frame quality and temporal coherence, making it ideal for video generation tasks. Additionally, we perform qualitative evaluation on the STAG4D dataset \cite{zeng2025stag4d}, generating 4D content for 28 more videos.


\textbf{Implementation Details.}
Our feature bank is implemented using Zero123++ v1.2~\cite{shi2023zero123++}, and we perform \(K=3\) iterations of generation. Our method does not require additional training; instead, we perform 75 inference steps per image. In the experiment, we set \(\tau = 0.98\) and \(\lambda = 0.8\) to ensure that fusion occurs only when local output features exhibit high similarity with feature bank features. This approach maintains consistency with previous frames while preserving information from the current frame.
\begin{table}[ht]
\centering
\footnotesize
\begin{tabular}{l c c c c}
\toprule
\textbf{Method} & \textbf{T-F} & \textbf{FVD (↓)} & \textbf{CLIP (↑)} & \textbf{LPIPS (↓)} \\
\midrule
SV4D \cite{xie2024sv4d}  & \(\times\) & 732.40 & 0.920 & 0.118 \\
4Diffusion \cite{zhang20244diffusion}\textsuperscript{+} & \(\times\) & 1551.63 & 0.873 & 0.228 \\
L4GM \cite{ren2024l4gm}\textsuperscript{+} & \(\times\) & 1360.04 & 0.913 & 0.158 \\
DS4D-GA \cite{yang2025not}\textsuperscript{+} & \(\times\) & 799.94 & 0.921 & 0.131 \\
DS4D-DA \cite{yang2025not}\textsuperscript{+} & \(\times\) & 784.02 & 0.923 & 0.131 \\
\midrule
Consistent4D \cite{jiang2023consistent4d}  & \checkmark & 1133.93 & 0.870 & 0.160 \\
4DGen \cite{yin20234dgen} & \checkmark & - & 0.894 & 0.130 \\
STAG4D \cite{zeng2025stag4d} & \checkmark & 992.21 & 0.909 & 0.126 \\
SC4D \cite{wu2024sc4d}\textsuperscript{+} & \checkmark & 852.98 & 0.912 & 0.137 \\
MVTokenFlow \cite{huang2025mvtokenflow}  & \checkmark & \underline{846.32} & \textbf{0.948} & \textbf{0.122} \\
\textbf{FB-4D (Ours)} & \checkmark & \textbf{724.26} & \underline{0.913} & \underline{0.125} \\
\bottomrule
\end{tabular}
\caption{\textbf{Comparison with state-of-the-art methods.} Our FB-4D achieves comparable performance with training-base method SV4D. T-F means training-free in Stage 1. (where \textsuperscript{+} indicates sourced from \cite{yang2025not})}
\label{table1}
\end{table}

\subsection{Experimental Results}

In this section, we conduct extensive and comprehensive comparisons with previously mentioned models, including training-based SV4D \cite{xie2024sv4d}, and training-free STAG4D \cite{zeng2025stag4d}, Consistent4D \cite{jiang2023consistent4d} and 4DGen \cite{yin20234dgen}. 
Our strong performance in both quantitative and qualitative evaluations validates the effectiveness of our design in improving both spatiotemporal consistency and the overall quality of 4D generation.

\begin{table}[ht]
\centering
\footnotesize
\begin{tabular}{c c | c c c}
\toprule
\textbf{FB} & \textbf{FB} & \textbf{FVD (↓)} & \textbf{CLIP (↑)} & \textbf{LPIPS (↓)} \\
{\textbf{(K \& V)}} & {\textbf{(Hidden States)}} &  &  &  \\
\midrule
\(\times\) & \(\times\) & 881.17 & 0.909 & 0.129 \\
\checkmark & \(\times\) & 832.16  & \textbf{0.911} & 0.127 \\
\(\times\) & \checkmark & 844.39  & 0.909 & 0.128 \\
\checkmark & \checkmark & \textbf{784.36} & 0.909 & \textbf{0.125} \\
\bottomrule
\end{tabular}
\caption{\textbf{Validation of the effectiveness of the feature bank.} FB means Feature Bank (only 1 iteration).}
\label{tab:featurebank_validation}
\end{table}

\begin{table}[ht]
\centering
\footnotesize
\begin{tabular}{c c c c}
\toprule
\textbf{Updating Method} & \textbf{FVD (↓)} & \textbf{CLIP (↑)} & \textbf{LPIPS (↓)} \\
\midrule
queue (1) & 818.24 & \textbf{0.910} & 0.128 \\
queue (2) & 844.48 & 0.908 & \textbf{0.125} \\
dynamic (Ours) & \textbf{784.36} & 0.909 & \textbf{0.125} \\
\bottomrule
\end{tabular}
\caption{\textbf{Comparison of different feature bank updating implementations} (only 1 iteration).}
\label{tab:queue}
\end{table}

\begin{table}[ht]
\centering
\footnotesize
\begin{tabular}{c c c c}
\toprule
\textbf{Layers}  & \textbf{FVD (↓)} & \textbf{CLIP (↑)} & \textbf{LPIPS (↓)} \\
\midrule
no layers & 881.17 & 0.909 & 0.128 \\
only upper layers & 865.43 & 0.905 & \textbf{0.125} \\
only lower layers &  852.82 & 0.910 & \textbf{0.125} \\
only middle layers & 821.44 & \textbf{0.911} & 0.127 \\
all layers (Ours) & \textbf{784.36} & 0.909 & \textbf{0.125} \\
\bottomrule
\end{tabular}
\caption{\textbf{Comparison of methods with and without featurebanks in different layers} (only 1 iteration).}
\label{tab:layers}
\end{table}

\noindent\textbf{Quantitative and Qualitative Results on Video-to-4D.}  
Table~\ref{table1} presents a comprehensive comparison between our method and several baselines across various evaluation metrics. Our approach outperforms all previous methods in video quality and smoothness, demonstrating superior realism and temporal consistency. Additionally, we achieve significantly better CLIP and LPIPS scores than Consistent4D, 4DGen, and STAG4D, indicating stronger semantic alignment with ground truth and enhanced realism. While SV4D surpasses our method on image-level metrics, our training-free design achieves comparable generation quality without requiring a 4D dataset.  

For qualitative evaluation, Figure~\ref{fig6} (a) visually compares our results with other methods on the Consistent4D dataset \cite{jiang2023consistent4d}, while Figure~\ref{fig6} (b) presents renderings from a dataset provided by STAG4D\cite{zeng2025stag4d}, with video sources from online resources. These comparisons further highlight the effectiveness of our approach in generating high-quality and temporally coherent 4D content.


\begin{table}[ht]
\centering
\footnotesize

\begin{tabular}{c c c c c}
\toprule
\textbf{P-F} & \textbf{P-I} & \textbf{FVD (↓)} & \textbf{CLIP (↑)} & \textbf{LPIPS (↓)} \\
\midrule
\(\times\) & \(\times\) & 902.28 & 0.909 & 0.128 \\
\checkmark & \(\times\) & 774.71  & 0.910 & \textbf{0.124} \\
\(\times\) & \checkmark & 805.74  & \textbf{0.914} & 0.125 \\
\checkmark & \checkmark & \textbf{758.41} & \textbf{0.914} & \textbf{0.124} \\
\bottomrule
\end{tabular}
\caption{\textbf{Validation of integrating past feature bank information into the iterative generation process.} P-F means previous feature banks and P-I means progressively iterations. (The iterations are controlled in 2 iters).}
\label{tab:multiple}
\end{table}



\subsection{Ablation Study}
\noindent\textbf{Feature Bank.} 
To validate the effectiveness of our feature bank in improving spatiotemporal consistency, we conducted an ablation study on two key innovations: key-value utilization and hidden states utilization. By limiting the iterations to one round, we controlled for variables and showed that our feature bank enhances both temporal and spatial consistency, improving generation quality. Results in Table \ref{tab:featurebank_validation} confirm that the feature bank boosts the fluidity and realism of the generated 4D content (Detailed visualization can be found in our Supplementary Materials)

We also tested different update strategies and the effect of inserting the feature bank at various layers. Results in Tables \ref{tab:queue} and \ref{tab:layers} indicate that inserting the feature bank at all layers improves performance. The queue update strategy limits past information retrieval, causing temporal inconsistencies, while increasing queue length without proper integration leads to redundancy and poor results. In contrast, our approach consolidates past information, improving spatiotemporal consistency (see supplementary materials for comparisons).

\noindent\textbf{Progressive Generation} 
This section demonstrates the effectiveness of progressive viewpoint selection and past feature integration over multiple iterations. Table~\ref{tab:multiple} shows that using past features enhances smoothness and realism. By selecting views closer to previous inputs, we improve feature utilization and generation quality. Visual results at different iterations are in the supplementary materials. Table~\ref{tab:last} confirms that three iterations yield effective results.

\begin{table}[ht]
\centering
\footnotesize
\begin{tabular}{c c c c}
\toprule
\textbf{Iterations} & \textbf{FVD (↓)} & \textbf{CLIP (↑)} & \textbf{LPIPS (↓)} \\
\midrule
1 & 784.36 & 0.909 & 0.125 \\
2 & 758.41 & \textbf{0.914} & \textbf{0.124} \\
3 & \textbf{724.26} & 0.913 & 0.125 \\
\bottomrule
\end{tabular}
\caption{\textbf{Comparison across different numbers of iterations.}}
\label{tab:last}
\end{table}

\section{Discussion and Conclusion}
We introduce FB-4D, a framework for dynamic 3D content generation from monocular videos. By utilizing feature banks, FB-4D improves spatial and temporal consistency, achieving high-quality 4D scene generation. Experimental results show FB-4D outperforms existing methods in rendering quality and consistency, with performance comparable to training-based approaches. It generates multi-view image sequences with an efficient vram cost of 14GB. Detailed computational cost information is available in the supplementary materials. Our approach sets a new benchmark for 4D consistency and rendering, with strong potential for real-world applications.
{
    \small
    \bibliographystyle{ieeenat_fullname}
    \bibliography{main}

\begin{thebibliography}{88}
\providecommand{\natexlab}[1]{#1}
\providecommand{\url}[1]{\texttt{#1}}
\expandafter\ifx\csname urlstyle\endcsname\relax
  \providecommand{\doi}[1]{doi: #1}\else
  \providecommand{\doi}{doi: \begingroup \urlstyle{rm}\Url}\fi

\bibitem[Bahmani et~al.(2024{\natexlab{a}})Bahmani, Liu, Yifan, Skorokhodov, Rong, Liu, Liu, Park, Tulyakov, Wetzstein, et~al.]{bahmani2024tc4d}
Sherwin Bahmani, Xian Liu, Wang Yifan, Ivan Skorokhodov, Victor Rong, Ziwei Liu, Xihui Liu, Jeong~Joon Park, Sergey Tulyakov, Gordon Wetzstein, et~al.
\newblock Tc4d: Trajectory-conditioned text-to-4d generation.
\newblock In \emph{European Conference on Computer Vision}, pages 53--72. Springer, 2024{\natexlab{a}}.

\bibitem[Bahmani et~al.(2024{\natexlab{b}})Bahmani, Skorokhodov, Rong, Wetzstein, Guibas, Wonka, Tulyakov, Park, Tagliasacchi, and Lindell]{bahmani20244d}
Sherwin Bahmani, Ivan Skorokhodov, Victor Rong, Gordon Wetzstein, Leonidas Guibas, Peter Wonka, Sergey Tulyakov, Jeong~Joon Park, Andrea Tagliasacchi, and David~B Lindell.
\newblock 4d-fy: Text-to-4d generation using hybrid score distillation sampling.
\newblock In \emph{Proceedings of the IEEE/CVF Conference on Computer Vision and Pattern Recognition}, pages 7996--8006, 2024{\natexlab{b}}.

\bibitem[Bahmani et~al.(2024{\natexlab{c}})Bahmani, Skorokhodov, Siarohin, Menapace, Qian, Vasilkovsky, Lee, Wang, Zou, Tagliasacchi, et~al.]{bahmani2024vd3d}
Sherwin Bahmani, Ivan Skorokhodov, Aliaksandr Siarohin, Willi Menapace, Guocheng Qian, Michael Vasilkovsky, Hsin-Ying Lee, Chaoyang Wang, Jiaxu Zou, Andrea Tagliasacchi, et~al.
\newblock Vd3d: Taming large video diffusion transformers for 3d camera control.
\newblock \emph{arXiv preprint arXiv:2407.12781}, 2024{\natexlab{c}}.

\bibitem[Biggs et~al.(2024)Biggs, Seshadri, Zou, Jain, Golatkar, Xie, Achille, Swaminathan, and Soatto]{biggs2024diffusion}
Benjamin Biggs, Arjun Seshadri, Yang Zou, Achin Jain, Aditya Golatkar, Yusheng Xie, Alessandro Achille, Ashwin Swaminathan, and Stefano Soatto.
\newblock Diffusion soup: Model merging for text-to-image diffusion models.
\newblock \emph{arXiv preprint arXiv:2406.08431}, 2024.

\bibitem[Blattmann et~al.(2023{\natexlab{a}})Blattmann, Dockhorn, Kulal, Mendelevitch, Kilian, Lorenz, Levi, English, Voleti, Letts, et~al.]{blattmann2023stable}
Andreas Blattmann, Tim Dockhorn, Sumith Kulal, Daniel Mendelevitch, Maciej Kilian, Dominik Lorenz, Yam Levi, Zion English, Vikram Voleti, Adam Letts, et~al.
\newblock Stable video diffusion: Scaling latent video diffusion models to large datasets.
\newblock \emph{arXiv preprint arXiv:2311.15127}, 2023{\natexlab{a}}.

\bibitem[Blattmann et~al.(2023{\natexlab{b}})Blattmann, Rombach, Ling, Dockhorn, Kim, Fidler, and Kreis]{blattmann2023align}
Andreas Blattmann, Robin Rombach, Huan Ling, Tim Dockhorn, Seung~Wook Kim, Sanja Fidler, and Karsten Kreis.
\newblock Align your latents: High-resolution video synthesis with latent diffusion models.
\newblock In \emph{Proceedings of the IEEE/CVF Conference on Computer Vision and Pattern Recognition}, pages 22563--22575, 2023{\natexlab{b}}.

\bibitem[Bolya et~al.(2022)Bolya, Fu, Dai, Zhang, Feichtenhofer, and Hoffman]{bolya2022token}
Daniel Bolya, Cheng-Yang Fu, Xiaoliang Dai, Peizhao Zhang, Christoph Feichtenhofer, and Judy Hoffman.
\newblock Token merging: Your vit but faster.
\newblock \emph{arXiv preprint arXiv:2210.09461}, 2022.

\bibitem[Chen et~al.(2024{\natexlab{a}})Chen, Chen, Ye, Gao, Chen, Fan, and Zhao]{chen2024ultraman}
Mingjin Chen, Junhao Chen, Xiaojun Ye, Huan-ang Gao, Xiaoxue Chen, Zhaoxin Fan, and Hao Zhao.
\newblock Ultraman: single image 3d human reconstruction with ultra speed and detail.
\newblock \emph{arXiv preprint arXiv:2403.12028}, 2024{\natexlab{a}}.

\bibitem[Chen et~al.(2024{\natexlab{b}})Chen, Wang, Wang, and Liu]{chen2024text}
Zilong Chen, Feng Wang, Yikai Wang, and Huaping Liu.
\newblock Text-to-3d using gaussian splatting.
\newblock In \emph{Proceedings of the IEEE/CVF Conference on Computer Vision and Pattern Recognition}, pages 21401--21412, 2024{\natexlab{b}}.

\bibitem[Chu et~al.(2025)Chu, Ke, and Fragkiadaki]{chu2025dreamscene4d}
Wen-Hsuan Chu, Lei Ke, and Katerina Fragkiadaki.
\newblock Dreamscene4d: Dynamic multi-object scene generation from monocular videos.
\newblock \emph{Advances in Neural Information Processing Systems}, 37:\penalty0 96181--96206, 2025.

\bibitem[Gao et~al.(2024)Gao, Gao, Li, Li, Zhi, Tang, and Zhao]{gao2024scp}
Huan-ang Gao, Mingju Gao, Jiaju Li, Wenyi Li, Rong Zhi, Hao Tang, and Hao Zhao.
\newblock Scp-diff: Spatial-categorical joint prior for diffusion based semantic image synthesis.
\newblock In \emph{European Conference on Computer Vision}, pages 37--54. Springer, 2024.

\bibitem[Gu{\'e}don and Lepetit(2024)]{guedon2024sugar}
Antoine Gu{\'e}don and Vincent Lepetit.
\newblock Sugar: Surface-aligned gaussian splatting for efficient 3d mesh reconstruction and high-quality mesh rendering.
\newblock In \emph{Proceedings of the IEEE/CVF Conference on Computer Vision and Pattern Recognition}, pages 5354--5363, 2024.

\bibitem[Guo et~al.(2023{\natexlab{a}})Guo, Hao, Caccavale, Ren, Zhang, Shan, Sankar, Schwing, Colburn, and Ma]{guo2023stabledreamer}
Pengsheng Guo, Hans Hao, Adam Caccavale, Zhongzheng Ren, Edward Zhang, Qi Shan, Aditya Sankar, Alexander~G Schwing, Alex Colburn, and Fangchang Ma.
\newblock Stabledreamer: Taming noisy score distillation sampling for text-to-3d.
\newblock \emph{arXiv preprint arXiv:2312.02189}, 2023{\natexlab{a}}.

\bibitem[Guo et~al.(2023{\natexlab{b}})Guo, Yang, Rao, Liang, Wang, Qiao, Agrawala, Lin, and Dai]{guo2023animatediff}
Yuwei Guo, Ceyuan Yang, Anyi Rao, Zhengyang Liang, Yaohui Wang, Yu Qiao, Maneesh Agrawala, Dahua Lin, and Bo Dai.
\newblock Animatediff: Animate your personalized text-to-image diffusion models without specific tuning.
\newblock \emph{arXiv preprint arXiv:2307.04725}, 2023{\natexlab{b}}.

\bibitem[Han et~al.(2024)Han, Gao, Kanazawa, Goel, and Gandelsman]{han2024more}
Xinyang Han, Zelin Gao, Angjoo Kanazawa, Shubham Goel, and Yossi Gandelsman.
\newblock The more you see in 2d the more you perceive in 3d.
\newblock In \emph{Proceedings of the IEEE/CVF Conference on Computer Vision and Pattern Recognition}, pages 20912--20922, 2024.

\bibitem[He et~al.(2024)He, Xu, Guo, Wetzstein, Dai, Li, and Yang]{he2024cameractrl}
Hao He, Yinghao Xu, Yuwei Guo, Gordon Wetzstein, Bo Dai, Hongsheng Li, and Ceyuan Yang.
\newblock Cameractrl: Enabling camera control for text-to-video generation.
\newblock \emph{arXiv preprint arXiv:2404.02101}, 2024.

\bibitem[He et~al.(2022)He, Yang, Zhang, Shan, and Chen]{he2022latent}
Yingqing He, Tianyu Yang, Yong Zhang, Ying Shan, and Qifeng Chen.
\newblock Latent video diffusion models for high-fidelity long video generation.
\newblock \emph{arXiv preprint arXiv:2211.13221}, 2022.

\bibitem[Ho et~al.(2022)Ho, Salimans, Gritsenko, Chan, Norouzi, and Fleet]{ho2022video}
Jonathan Ho, Tim Salimans, Alexey Gritsenko, William Chan, Mohammad Norouzi, and David~J Fleet.
\newblock Video diffusion models.
\newblock \emph{Advances in Neural Information Processing Systems}, 35:\penalty0 8633--8646, 2022.

\bibitem[Hong et~al.(2023)Hong, Zhang, Gu, Bi, Zhou, Liu, Liu, Sunkavalli, Bui, and Tan]{hong2023lrm}
Yicong Hong, Kai Zhang, Jiuxiang Gu, Sai Bi, Yang Zhou, Difan Liu, Feng Liu, Kalyan Sunkavalli, Trung Bui, and Hao Tan.
\newblock Lrm: Large reconstruction model for single image to 3d.
\newblock \emph{arXiv preprint arXiv:2311.04400}, 2023.

\bibitem[Huang et~al.(2025)Huang, Liu, Zheng, Wang, Dou, and Yang]{huang2025mvtokenflow}
Hanzhuo Huang, Yuan Liu, Ge Zheng, Jiepeng Wang, Zhiyang Dou, and Sibei Yang.
\newblock Mvtokenflow: High-quality 4d content generation using multiview token flow.
\newblock \emph{arXiv preprint arXiv:2502.11697}, 2025.

\bibitem[Jiang et~al.(2023{\natexlab{a}})Jiang, Jiang, Zhao, and Huang]{jiang2023leap}
Hanwen Jiang, Zhenyu Jiang, Yue Zhao, and Qixing Huang.
\newblock Leap: Liberate sparse-view 3d modeling from camera poses.
\newblock \emph{arXiv preprint arXiv:2310.01410}, 2023{\natexlab{a}}.

\bibitem[Jiang et~al.(2023{\natexlab{b}})Jiang, Zhang, Gao, Hu, and Yao]{jiang2023consistent4d}
Yanqin Jiang, Li Zhang, Jin Gao, Weimin Hu, and Yao Yao.
\newblock Consistent4d: Consistent 360 $\{$$\backslash$deg$\}$ dynamic object generation from monocular video.
\newblock \emph{arXiv preprint arXiv:2311.02848}, 2023{\natexlab{b}}.

\bibitem[Jiang et~al.(2025)Jiang, Yu, Cao, Wang, Hu, and Gao]{jiang2025animate3d}
Yanqin Jiang, Chaohui Yu, Chenjie Cao, Fan Wang, Weiming Hu, and Jin Gao.
\newblock Animate3d: Animating any 3d model with multi-view video diffusion.
\newblock \emph{Advances in Neural Information Processing Systems}, 37:\penalty0 125879--125906, 2025.

\bibitem[Karnewar et~al.(2023)Karnewar, Mitra, Vedaldi, and Novotny]{karnewar2023holofusion}
Animesh Karnewar, Niloy~J Mitra, Andrea Vedaldi, and David Novotny.
\newblock Holofusion: Towards photo-realistic 3d generative modeling.
\newblock In \emph{Proceedings of the IEEE/CVF International Conference on Computer Vision}, pages 22976--22985, 2023.

\bibitem[Kerbl et~al.(2023)Kerbl, Kopanas, Leimkuehler, and Drettakis]{kerbl20233dgs}
Bernhard Kerbl, Georgios Kopanas, Thomas Leimkuehler, and George Drettakis.
\newblock 3d gaussian splatting for real-time radiance field rendering.
\newblock \emph{ACM Transactions on Graphics (TOG)}, 42\penalty0 (4):\penalty0 139:1 -- 139:14, 2023.

\bibitem[Li et~al.(2025)Li, Zhou, Liu, Wang, Zhuang, Gao, Jin, and Zhao]{li2025avd2}
Cheng Li, Keyuan Zhou, Tong Liu, Yu Wang, Mingqiao Zhuang, Huan-ang Gao, Bu Jin, and Hao Zhao.
\newblock Avd2: Accident video diffusion for accident video description.
\newblock \emph{arXiv preprint arXiv:2502.14801}, 2025.

\bibitem[Li et~al.(2023{\natexlab{a}})Li, Tan, Zhang, Xu, Luan, Xu, Hong, Sunkavalli, Shakhnarovich, and Bi]{li2023instant3d}
Jiahao Li, Hao Tan, Kai Zhang, Zexiang Xu, Fujun Luan, Yinghao Xu, Yicong Hong, Kalyan Sunkavalli, Greg Shakhnarovich, and Sai Bi.
\newblock Instant3d: Fast text-to-3d with sparse-view generation and large reconstruction model.
\newblock \emph{arXiv preprint arXiv:2311.06214}, 2023{\natexlab{a}}.

\bibitem[Li et~al.(2024{\natexlab{a}})Li, Liu, Long, Zhang, Lin, Li, Qi, Zhang, Xue, Luo, et~al.]{li2024era3d}
Peng Li, Yuan Liu, Xiaoxiao Long, Feihu Zhang, Cheng Lin, Mengfei Li, Xingqun Qi, Shanghang Zhang, Wei Xue, Wenhan Luo, et~al.
\newblock Era3d: high-resolution multiview diffusion using efficient row-wise attention.
\newblock \emph{Advances in Neural Information Processing Systems}, 37:\penalty0 55975--56000, 2024{\natexlab{a}}.

\bibitem[Li et~al.(2024{\natexlab{b}})Li, Pan, Yang, Xu, Zhou, Zhang, Li, Kadambi, Wang, Tu, et~al.]{li20244k4dgen}
Renjie Li, Panwang Pan, Bangbang Yang, Dejia Xu, Shijie Zhou, Xuanyang Zhang, Zeming Li, Achuta Kadambi, Zhangyang Wang, Zhengzhong Tu, et~al.
\newblock 4k4dgen: Panoramic 4d generation at 4k resolution.
\newblock \emph{arXiv preprint arXiv:2406.13527}, 2024{\natexlab{b}}.

\bibitem[Li et~al.(2023{\natexlab{b}})Li, Chen, Chen, and Tan]{li2023sweetdreamer}
Weiyu Li, Rui Chen, Xuelin Chen, and Ping Tan.
\newblock Sweetdreamer: Aligning geometric priors in 2d diffusion for consistent text-to-3d.
\newblock \emph{arXiv preprint arXiv:2310.02596}, 2023{\natexlab{b}}.

\bibitem[Li et~al.(2024{\natexlab{c}})Li, Gao, Gao, Tian, Zhi, and Zhao]{li2024training}
Wenyi Li, Huan-ang Gao, Mingju Gao, Beiwen Tian, Rong Zhi, and Hao Zhao.
\newblock Training-free model merging for multi-target domain adaptation.
\newblock \emph{arXiv preprint arXiv:2407.13771}, 2024{\natexlab{c}}.

\bibitem[Li et~al.(2024{\natexlab{d}})Li, Xu, Zhang, Gao, Gao, Wang, and Zhao]{li2024fairdiff}
Wenyi Li, Haoran Xu, Guiyu Zhang, Huan-ang Gao, Mingju Gao, Mengyu Wang, and Hao Zhao.
\newblock Fairdiff: Fair segmentation with point-image diffusion.
\newblock In \emph{International Conference on Medical Image Computing and Computer-Assisted Intervention}, pages 617--628. Springer, 2024{\natexlab{d}}.

\bibitem[Liang et~al.(2024)Liang, Yang, Lin, Li, Xu, and Chen]{liang2024luciddreamer}
Yixun Liang, Xin Yang, Jiantao Lin, Haodong Li, Xiaogang Xu, and Yingcong Chen.
\newblock Luciddreamer: Towards high-fidelity text-to-3d generation via interval score matching.
\newblock In \emph{Proceedings of the IEEE/CVF Conference on Computer Vision and Pattern Recognition}, pages 6517--6526, 2024.

\bibitem[Ling et~al.(2024)Ling, Kim, Torralba, Fidler, and Kreis]{ling2024align}
Huan Ling, Seung~Wook Kim, Antonio Torralba, Sanja Fidler, and Karsten Kreis.
\newblock Align your gaussians: Text-to-4d with dynamic 3d gaussians and composed diffusion models.
\newblock In \emph{Proceedings of the IEEE/CVF Conference on Computer Vision and Pattern Recognition}, pages 8576--8588, 2024.

\bibitem[Liu et~al.(2024{\natexlab{a}})Liu, Hu, Yang, Chen, Wang, Chen, Cai, Gao, and Zhao]{liu2024rip}
Junchen Liu, Wenbo Hu, Zhuo Yang, Jianteng Chen, Guoliang Wang, Xiaoxue Chen, Yantong Cai, Huan-ang Gao, and Hao Zhao.
\newblock Rip-nerf: Anti-aliasing radiance fields with ripmap-encoded platonic solids.
\newblock In \emph{ACM SIGGRAPH 2024 Conference Papers}, pages 1--11, 2024{\natexlab{a}}.

\bibitem[Liu et~al.(2024{\natexlab{b}})Liu, Shi, Chen, Zhang, Xu, Wei, Chen, Zeng, Gu, and Su]{liu2024one_a}
Minghua Liu, Ruoxi Shi, Linghao Chen, Zhuoyang Zhang, Chao Xu, Xinyue Wei, Hansheng Chen, Chong Zeng, Jiayuan Gu, and Hao Su.
\newblock One-2-3-45++: Fast single image to 3d objects with consistent multi-view generation and 3d diffusion.
\newblock In \emph{Proceedings of the IEEE/CVF Conference on Computer Vision and Pattern Recognition}, pages 10072--10083, 2024{\natexlab{b}}.

\bibitem[Liu et~al.(2024{\natexlab{c}})Liu, Xu, Jin, Chen, Varma~T, Xu, and Su]{liu2024one_b}
Minghua Liu, Chao Xu, Haian Jin, Linghao Chen, Mukund Varma~T, Zexiang Xu, and Hao Su.
\newblock One-2-3-45: Any single image to 3d mesh in 45 seconds without per-shape optimization.
\newblock \emph{Advances in Neural Information Processing Systems}, 36, 2024{\natexlab{c}}.

\bibitem[Liu et~al.(2023{\natexlab{a}})Liu, Wu, Van~Hoorick, Tokmakov, Zakharov, and Vondrick]{liu2023zero}
Ruoshi Liu, Rundi Wu, Basile Van~Hoorick, Pavel Tokmakov, Sergey Zakharov, and Carl Vondrick.
\newblock Zero-1-to-3: Zero-shot one image to 3d object.
\newblock In \emph{Proceedings of the IEEE/CVF international conference on computer vision}, pages 9298--9309, 2023{\natexlab{a}}.

\bibitem[Liu et~al.(2023{\natexlab{b}})Liu, Lin, Zeng, Long, Liu, Komura, and Wang]{liu2023syncdreamer}
Yuan Liu, Cheng Lin, Zijiao Zeng, Xiaoxiao Long, Lingjie Liu, Taku Komura, and Wenping Wang.
\newblock Syncdreamer: Generating multiview-consistent images from a single-view image.
\newblock \emph{arXiv preprint arXiv:2309.03453}, 2023{\natexlab{b}}.

\bibitem[Long et~al.(2024)Long, Guo, Lin, Liu, Dou, Liu, Ma, Zhang, Habermann, Theobalt, et~al.]{long2024wonder3d}
Xiaoxiao Long, Yuan-Chen Guo, Cheng Lin, Yuan Liu, Zhiyang Dou, Lingjie Liu, Yuexin Ma, Song-Hai Zhang, Marc Habermann, Christian Theobalt, et~al.
\newblock Wonder3d: Single image to 3d using cross-domain diffusion.
\newblock In \emph{Proceedings of the IEEE/CVF Conference on Computer Vision and Pattern Recognition}, pages 9970--9980, 2024.

\bibitem[Luo et~al.(2024)Luo, Dunlap, Park, Holynski, and Darrell]{luo2024diffusion}
Grace Luo, Lisa Dunlap, Dong~Huk Park, Aleksander Holynski, and Trevor Darrell.
\newblock Diffusion hyperfeatures: Searching through time and space for semantic correspondence.
\newblock \emph{Advances in Neural Information Processing Systems}, 36, 2024.

\bibitem[M{\"u}ller et~al.(2022)M{\"u}ller, Evans, Schied, and Keller]{muller2022instant}
Thomas M{\"u}ller, Alex Evans, Christoph Schied, and Alexander Keller.
\newblock Instant neural graphics primitives with a multiresolution hash encoding.
\newblock \emph{ACM Transactions on Graphics (ToG)}, 41\penalty0 (4):\penalty0 1--15, 2022.

\bibitem[Ni et~al.(2025)Ni, Feng, Chi, Zheng, Gao, Ma, Ma, and Lan]{ni2025straight}
Yuyan Ni, Shikun Feng, Haohan Chi, Bowen Zheng, Huan-ang Gao, Wei-Ying Ma, Zhi-Ming Ma, and Yanyan Lan.
\newblock Straight-line diffusion model for efficient 3d molecular generation.
\newblock \emph{arXiv preprint arXiv:2503.02918}, 2025.

\bibitem[Ren et~al.(2023)Ren, Pan, Tang, Zhang, Cao, Zeng, and Liu]{ren2023dreamgaussian4d}
Jiawei Ren, Liang Pan, Jiaxiang Tang, Chi Zhang, Ang Cao, Gang Zeng, and Ziwei Liu.
\newblock Dreamgaussian4d: Generative 4d gaussian splatting.
\newblock \emph{arXiv preprint arXiv:2312.17142}, 2023.

\bibitem[Ren et~al.(2024)Ren, Xie, Mirzaei, Kreis, Liu, Torralba, Fidler, Kim, Ling, et~al.]{ren2024l4gm}
Jiawei Ren, Cheng Xie, Ashkan Mirzaei, Karsten Kreis, Ziwei Liu, Antonio Torralba, Sanja Fidler, Seung~Wook Kim, Huan Ling, et~al.
\newblock L4gm: Large 4d gaussian reconstruction model.
\newblock \emph{Advances in Neural Information Processing Systems}, 37:\penalty0 56828--56858, 2024.

\bibitem[Rombach et~al.(2022)Rombach, Blattmann, Lorenz, Esser, and Ommer]{rombach2022high}
Robin Rombach, Andreas Blattmann, Dominik Lorenz, Patrick Esser, and Bj{\"o}rn Ommer.
\newblock High-resolution image synthesis with latent diffusion models.
\newblock In \emph{Proceedings of the IEEE/CVF conference on computer vision and pattern recognition}, pages 10684--10695, 2022.

\bibitem[Sargent et~al.(2023)Sargent, Li, Shah, Herrmann, Yu, Zhang, Chan, Lagun, Fei-Fei, Sun, et~al.]{sargent2023zeronvs}
Kyle Sargent, Zizhang Li, Tanmay Shah, Charles Herrmann, Hong-Xing Yu, Yunzhi Zhang, Eric~Ryan Chan, Dmitry Lagun, Li Fei-Fei, Deqing Sun, et~al.
\newblock Zeronvs: Zero-shot 360-degree view synthesis from a single real image.
\newblock \emph{arXiv preprint arXiv:2310.17994}, 2023.

\bibitem[Shi et~al.(2023{\natexlab{a}})Shi, Chen, Zhang, Liu, Xu, Wei, Chen, Zeng, and Su]{shi2023zero123++}
Ruoxi Shi, Hansheng Chen, Zhuoyang Zhang, Minghua Liu, Chao Xu, Xinyue Wei, Linghao Chen, Chong Zeng, and Hao Su.
\newblock Zero123++: a single image to consistent multi-view diffusion base model.
\newblock \emph{arXiv preprint arXiv:2310.15110}, 2023{\natexlab{a}}.

\bibitem[Shi et~al.(2023{\natexlab{b}})Shi, Wang, Cao, Tang, Qi, Yang, Huang, Liu, Zhang, and Shum]{shi2023toss}
Yukai Shi, Jianan Wang, He Cao, Boshi Tang, Xianbiao Qi, Tianyu Yang, Yukun Huang, Shilong Liu, Lei Zhang, and Heung-Yeung Shum.
\newblock Toss: High-quality text-guided novel view synthesis from a single image.
\newblock \emph{arXiv preprint arXiv:2310.10644}, 2023{\natexlab{b}}.

\bibitem[Shi et~al.(2023{\natexlab{c}})Shi, Wang, Ye, Long, Li, and Yang]{shi2023mvdream}
Yichun Shi, Peng Wang, Jianglong Ye, Mai Long, Kejie Li, and Xiao Yang.
\newblock Mvdream: Multi-view diffusion for 3d generation.
\newblock \emph{arXiv preprint arXiv:2308.16512}, 2023{\natexlab{c}}.

\bibitem[Singer et~al.(2022)Singer, Polyak, Hayes, Yin, An, Zhang, Hu, Yang, Ashual, Gafni, et~al.]{singer2022make}
Uriel Singer, Adam Polyak, Thomas Hayes, Xi Yin, Jie An, Songyang Zhang, Qiyuan Hu, Harry Yang, Oron Ashual, Oran Gafni, et~al.
\newblock Make-a-video: Text-to-video generation without text-video data.
\newblock \emph{arXiv preprint arXiv:2209.14792}, 2022.

\bibitem[Singer et~al.(2023)Singer, Sheynin, Polyak, Ashual, Makarov, Kokkinos, Goyal, Vedaldi, Parikh, Johnson, et~al.]{singer2023text}
Uriel Singer, Shelly Sheynin, Adam Polyak, Oron Ashual, Iurii Makarov, Filippos Kokkinos, Naman Goyal, Andrea Vedaldi, Devi Parikh, Justin Johnson, et~al.
\newblock Text-to-4d dynamic scene generation.
\newblock \emph{arXiv preprint arXiv:2301.11280}, 2023.

\bibitem[Sun et~al.(2023)Sun, Zhang, Shao, Wang, Liu, Xie, and Liu]{sun2023dreamcraft3d}
Jingxiang Sun, Bo Zhang, Ruizhi Shao, Lizhen Wang, Wen Liu, Zhenda Xie, and Yebin Liu.
\newblock Dreamcraft3d: Hierarchical 3d generation with bootstrapped diffusion prior.
\newblock \emph{arXiv preprint arXiv:2310.16818}, 2023.

\bibitem[Sun et~al.(2024)Sun, Guo, Wan, Yan, Yin, Zhou, Liao, and Li]{sun2024eg4d}
Qi Sun, Zhiyang Guo, Ziyu Wan, Jing~Nathan Yan, Shengming Yin, Wengang Zhou, Jing Liao, and Houqiang Li.
\newblock Eg4d: Explicit generation of 4d object without score distillation.
\newblock \emph{arXiv preprint arXiv:2405.18132}, 2024.

\bibitem[Tang et~al.(2023{\natexlab{a}})Tang, Ren, Zhou, Liu, and Zeng]{tang2023dreamgaussian}
Jiaxiang Tang, Jiawei Ren, Hang Zhou, Ziwei Liu, and Gang Zeng.
\newblock Dreamgaussian: Generative gaussian splatting for efficient 3d content creation.
\newblock \emph{arXiv preprint arXiv:2309.16653}, 2023{\natexlab{a}}.

\bibitem[Tang et~al.(2023{\natexlab{b}})Tang, Jia, Wang, Phoo, and Hariharan]{tang2023emergent}
Luming Tang, Menglin Jia, Qianqian Wang, Cheng~Perng Phoo, and Bharath Hariharan.
\newblock Emergent correspondence from image diffusion.
\newblock \emph{Advances in Neural Information Processing Systems}, 36:\penalty0 1363--1389, 2023{\natexlab{b}}.

\bibitem[Tochilkin et~al.(2024)Tochilkin, Pankratz, Liu, Huang, Letts, Li, Liang, Laforte, Jampani, and Cao]{tochilkin2024triposr}
Dmitry Tochilkin, David Pankratz, Zexiang Liu, Zixuan Huang, Adam Letts, Yangguang Li, Ding Liang, Christian Laforte, Varun Jampani, and Yan-Pei Cao.
\newblock Triposr: Fast 3d object reconstruction from a single image.
\newblock \emph{arXiv preprint arXiv:2403.02151}, 2024.

\bibitem[Voleti et~al.(2022)Voleti, Jolicoeur-Martineau, and Pal]{voleti2022mcvd}
Vikram Voleti, Alexia Jolicoeur-Martineau, and Chris Pal.
\newblock Mcvd-masked conditional video diffusion for prediction, generation, and interpolation.
\newblock \emph{Advances in neural information processing systems}, 35:\penalty0 23371--23385, 2022.

\bibitem[Voleti et~al.(2025)Voleti, Yao, Boss, Letts, Pankratz, Tochilkin, Laforte, Rombach, and Jampani]{voleti2025sv3d}
Vikram Voleti, Chun-Han Yao, Mark Boss, Adam Letts, David Pankratz, Dmitry Tochilkin, Christian Laforte, Robin Rombach, and Varun Jampani.
\newblock Sv3d: Novel multi-view synthesis and 3d generation from a single image using latent video diffusion.
\newblock In \emph{European Conference on Computer Vision}, pages 439--457. Springer, 2025.

\bibitem[Wang et~al.(2023{\natexlab{a}})Wang, Yuan, Chen, Zhang, Wang, and Zhang]{wang2023modelscope}
Jiuniu Wang, Hangjie Yuan, Dayou Chen, Yingya Zhang, Xiang Wang, and Shiwei Zhang.
\newblock Modelscope text-to-video technical report.
\newblock \emph{arXiv preprint arXiv:2308.06571}, 2023{\natexlab{a}}.

\bibitem[Wang and Shi(2023)]{wang2023imagedream}
Peng Wang and Yichun Shi.
\newblock Imagedream: Image-prompt multi-view diffusion for 3d generation.
\newblock \emph{arXiv preprint arXiv:2312.02201}, 2023.

\bibitem[Wang et~al.(2023{\natexlab{b}})Wang, Tan, Bi, Xu, Luan, Sunkavalli, Wang, Xu, and Zhang]{wang2023pf}
Peng Wang, Hao Tan, Sai Bi, Yinghao Xu, Fujun Luan, Kalyan Sunkavalli, Wenping Wang, Zexiang Xu, and Kai Zhang.
\newblock Pf-lrm: Pose-free large reconstruction model for joint pose and shape prediction.
\newblock \emph{arXiv preprint arXiv:2311.12024}, 2023{\natexlab{b}}.

\bibitem[Wang et~al.(2024{\natexlab{a}})Wang, Lu, Wang, Bao, Li, Su, and Zhu]{wang2024prolificdreamer}
Zhengyi Wang, Cheng Lu, Yikai Wang, Fan Bao, Chongxuan Li, Hang Su, and Jun Zhu.
\newblock Prolificdreamer: High-fidelity and diverse text-to-3d generation with variational score distillation.
\newblock \emph{Advances in Neural Information Processing Systems}, 36, 2024{\natexlab{a}}.

\bibitem[Wang et~al.(2024{\natexlab{b}})Wang, Yuan, Wang, Li, Chen, Xia, Luo, and Shan]{wang2024motionctrl}
Zhouxia Wang, Ziyang Yuan, Xintao Wang, Yaowei Li, Tianshui Chen, Menghan Xia, Ping Luo, and Ying Shan.
\newblock Motionctrl: A unified and flexible motion controller for video generation.
\newblock In \emph{ACM SIGGRAPH 2024 Conference Papers}, pages 1--11, 2024{\natexlab{b}}.

\bibitem[Wei et~al.(2024)Wei, Zhang, Bi, Tan, Luan, Deschaintre, Sunkavalli, Su, and Xu]{wei2024meshlrm}
Xinyue Wei, Kai Zhang, Sai Bi, Hao Tan, Fujun Luan, Valentin Deschaintre, Kalyan Sunkavalli, Hao Su, and Zexiang Xu.
\newblock Meshlrm: Large reconstruction model for high-quality mesh.
\newblock \emph{arXiv preprint arXiv:2404.12385}, 2024.

\bibitem[Weng et~al.(2023)Weng, Yang, Wang, Li, Zhang, Chen, and Zhang]{weng2023consistent123}
Haohan Weng, Tianyu Yang, Jianan Wang, Yu Li, Tong Zhang, CL Chen, and Lei Zhang.
\newblock Consistent123: Improve consistency for one image to 3d object synthesis.
\newblock \emph{arXiv preprint arXiv:2310.08092}, 2023.

\bibitem[Wu et~al.(2024)Wu, Yu, Jiang, Cao, Wang, and Bai]{wu2024sc4d}
Zijie Wu, Chaohui Yu, Yanqin Jiang, Chenjie Cao, Fan Wang, and Xiang Bai.
\newblock Sc4d: Sparse-controlled video-to-4d generation and motion transfer.
\newblock In \emph{European Conference on Computer Vision}, pages 361--379. Springer, 2024.

\bibitem[Wu et~al.(2025)Wu, Yu, Jiang, Cao, Wang, and Bai]{wu2025sc4d}
Zijie Wu, Chaohui Yu, Yanqin Jiang, Chenjie Cao, Fan Wang, and Xiang Bai.
\newblock Sc4d: Sparse-controlled video-to-4d generation and motion transfer.
\newblock In \emph{European Conference on Computer Vision}, pages 361--379. Springer, 2025.

\bibitem[Xie et~al.(2024{\natexlab{a}})Xie, Zong, Qiu, Li, Feng, Yang, and Jiang]{xie2024physgaussian}
Tianyi Xie, Zeshun Zong, Yuxing Qiu, Xuan Li, Yutao Feng, Yin Yang, and Chenfanfu Jiang.
\newblock Physgaussian: Physics-integrated 3d gaussians for generative dynamics.
\newblock In \emph{Proceedings of the IEEE/CVF Conference on Computer Vision and Pattern Recognition}, pages 4389--4398, 2024{\natexlab{a}}.

\bibitem[Xie et~al.(2024{\natexlab{b}})Xie, Yao, Voleti, Jiang, and Jampani]{xie2024sv4d}
Yiming Xie, Chun-Han Yao, Vikram Voleti, Huaizu Jiang, and Varun Jampani.
\newblock Sv4d: Dynamic 3d content generation with multi-frame and multi-view consistency.
\newblock \emph{arXiv preprint arXiv:2407.17470}, 2024{\natexlab{b}}.

\bibitem[Xu et~al.(2022)Xu, Xu, Philip, Bi, Shu, Sunkavalli, and Neumann]{xu2022point}
Qiangeng Xu, Zexiang Xu, Julien Philip, Sai Bi, Zhixin Shu, Kalyan Sunkavalli, and Ulrich Neumann.
\newblock Point-nerf: Point-based neural radiance fields.
\newblock In \emph{Proceedings of the IEEE/CVF conference on computer vision and pattern recognition}, pages 5438--5448, 2022.

\bibitem[Xu et~al.(2024)Xu, Chen, Gao, Zhao, Zhang, and Zhao]{xu2024diffusion}
Zhiyuan Xu, Yinhe Chen, Huan-ang Gao, Weiyan Zhao, Guiyu Zhang, and Hao Zhao.
\newblock Diffusion-based visual anagram as multi-task learning.
\newblock \emph{arXiv preprint arXiv:2412.02693}, 2024.

\bibitem[Yang et~al.(2024{\natexlab{a}})Yang, Shen, Guo, Wang, Cao, Zhang, and Tao]{yang2024model}
Enneng Yang, Li Shen, Guibing Guo, Xingwei Wang, Xiaochun Cao, Jie Zhang, and Dacheng Tao.
\newblock Model merging in llms, mllms, and beyond: Methods, theories, applications and opportunities.
\newblock \emph{arXiv preprint arXiv:2408.07666}, 2024{\natexlab{a}}.

\bibitem[Yang et~al.(2025)Yang, Liu, Zhu, Liu, Ma, Nong, and Liang]{yang2025not}
Liying Yang, Chen Liu, Zhenwei Zhu, Ajian Liu, Hui Ma, Jian Nong, and Yanyan Liang.
\newblock Not all frame features are equal: Video-to-4d generation via decoupling dynamic-static features.
\newblock \emph{arXiv preprint arXiv:2502.08377}, 2025.

\bibitem[Yang et~al.(2024{\natexlab{b}})Yang, Pan, Gu, and Zhang]{yang2024diffusion}
Zeyu Yang, Zijie Pan, Chun Gu, and Li Zhang.
\newblock Diffusion$^2$: Dynamic 3d content generation via score composition of orthogonal diffusion models.
\newblock \emph{arXiv preprint arXiv:2404.02148}, 2024{\natexlab{b}}.

\bibitem[Ye et~al.(2024)Ye, Wang, Li, Shi, and Wang]{ye2024consistent}
Jianglong Ye, Peng Wang, Kejie Li, Yichun Shi, and Heng Wang.
\newblock Consistent-1-to-3: Consistent image to 3d view synthesis via geometry-aware diffusion models.
\newblock In \emph{2024 International Conference on 3D Vision (3DV)}, pages 664--674. IEEE, 2024.

\bibitem[Yi et~al.(2023)Yi, Fang, Wu, Xie, Zhang, Liu, Tian, and Wang]{yi2023gaussiandreamer}
Taoran Yi, Jiemin Fang, Guanjun Wu, Lingxi Xie, Xiaopeng Zhang, Wenyu Liu, Qi Tian, and Xinggang Wang.
\newblock Gaussiandreamer: Fast generation from text to 3d gaussian splatting with point cloud priors.
\newblock \emph{arXiv preprint arXiv:2310.08529}, 2023.

\bibitem[Yin et~al.(2023)Yin, Xu, Wang, Zhao, and Wei]{yin20234dgen}
Yuyang Yin, Dejia Xu, Zhangyang Wang, Yao Zhao, and Yunchao Wei.
\newblock 4dgen: Grounded 4d content generation with spatial-temporal consistency.
\newblock \emph{arXiv preprint arXiv:2312.17225}, 2023.

\bibitem[Yu et~al.(2024{\natexlab{a}})Yu, Wang, Zhuang, Menapace, Siarohin, Cao, Jeni, Tulyakov, and Lee]{yu20244real}
Heng Yu, Chaoyang Wang, Peiye Zhuang, Willi Menapace, Aliaksandr Siarohin, Junli Cao, L{\'a}szl{\'o} Jeni, Sergey Tulyakov, and Hsin-Ying Lee.
\newblock 4real: Towards photorealistic 4d scene generation via video diffusion models.
\newblock \emph{Advances in Neural Information Processing Systems}, 37:\penalty0 45256--45280, 2024{\natexlab{a}}.

\bibitem[Yu et~al.(2024{\natexlab{b}})Yu, Chen, Huang, Sattler, and Geiger]{yu2024mip}
Zehao Yu, Anpei Chen, Binbin Huang, Torsten Sattler, and Andreas Geiger.
\newblock Mip-splatting: Alias-free 3d gaussian splatting.
\newblock In \emph{Proceedings of the IEEE/CVF Conference on Computer Vision and Pattern Recognition}, pages 19447--19456, 2024{\natexlab{b}}.

\bibitem[Zeng et~al.(2025)Zeng, Jiang, Zhu, Lu, Lin, Zhu, Hu, Cao, and Yao]{zeng2025stag4d}
Yifei Zeng, Yanqin Jiang, Siyu Zhu, Yuanxun Lu, Youtian Lin, Hao Zhu, Weiming Hu, Xun Cao, and Yao Yao.
\newblock Stag4d: Spatial-temporal anchored generative 4d gaussians.
\newblock In \emph{European Conference on Computer Vision}, pages 163--179. Springer, 2025.

\bibitem[Zhang et~al.(2024{\natexlab{a}})Zhang, Gao, Jiang, Zhao, and Zheng]{zhang2024ctrl}
Guiyu Zhang, Huan-ang Gao, Zijian Jiang, Hao Zhao, and Zhedong Zheng.
\newblock Ctrl-u: Robust conditional image generation via uncertainty-aware reward modeling.
\newblock \emph{arXiv preprint arXiv:2410.11236}, 2024{\natexlab{a}}.

\bibitem[Zhang et~al.(2024{\natexlab{b}})Zhang, Chen, Wang, Liu, Wang, and Qiao]{zhang20244diffusion}
Haiyu Zhang, Xinyuan Chen, Yaohui Wang, Xihui Liu, Yunhong Wang, and Yu Qiao.
\newblock 4diffusion: Multi-view video diffusion model for 4d generation.
\newblock \emph{Advances in Neural Information Processing Systems}, 37:\penalty0 15272--15295, 2024{\natexlab{b}}.

\bibitem[Zhang et~al.(2023)Zhang, Rao, and Agrawala]{zhang2023adding}
Lvmin Zhang, Anyi Rao, and Maneesh Agrawala.
\newblock Adding conditional control to text-to-image diffusion models.
\newblock In \emph{Proceedings of the IEEE/CVF international conference on computer vision}, pages 3836--3847, 2023.

\bibitem[Zhao et~al.(2023)Zhao, Yan, Xie, Hong, Li, and Lee]{zhao2023animate124}
Yuyang Zhao, Zhiwen Yan, Enze Xie, Lanqing Hong, Zhenguo Li, and Gim~Hee Lee.
\newblock Animate124: Animating one image to 4d dynamic scene.
\newblock \emph{arXiv preprint arXiv:2311.14603}, 2023.

\bibitem[Zhao et~al.(2024)Zhao, Lin, Lin, Yan, Li, Yang, Wang, Lee, and Wang]{zhao2024genxd}
Yuyang Zhao, Chung-Ching Lin, Kevin Lin, Zhiwen Yan, Linjie Li, Zhengyuan Yang, Jianfeng Wang, Gim~Hee Lee, and Lijuan Wang.
\newblock Genxd: Generating any 3d and 4d scenes.
\newblock \emph{arXiv preprint arXiv:2411.02319}, 2024.

\bibitem[Zhou et~al.(2024)Zhou, Shih, Meng, and Ermon]{zhou2024dreampropeller}
Linqi Zhou, Andy Shih, Chenlin Meng, and Stefano Ermon.
\newblock Dreampropeller: Supercharge text-to-3d generation with parallel sampling.
\newblock In \emph{Proceedings of the IEEE/CVF Conference on Computer Vision and Pattern Recognition}, pages 4610--4619, 2024.

\bibitem[Zou et~al.(2024)Zou, Yu, Guo, Li, Liang, Cao, and Zhang]{zou2024triplane}
Zi-Xin Zou, Zhipeng Yu, Yuan-Chen Guo, Yangguang Li, Ding Liang, Yan-Pei Cao, and Song-Hai Zhang.
\newblock Triplane meets gaussian splatting: Fast and generalizable single-view 3d reconstruction with transformers.
\newblock In \emph{Proceedings of the IEEE/CVF Conference on Computer Vision and Pattern Recognition}, pages 10324--10335, 2024.

\end{thebibliography}
}

\appendix 
\section{More Experiments}

\begin{figure}[ht]     
    \centering     
    \includegraphics[width=0.6\linewidth, height = 0.85\linewidth]{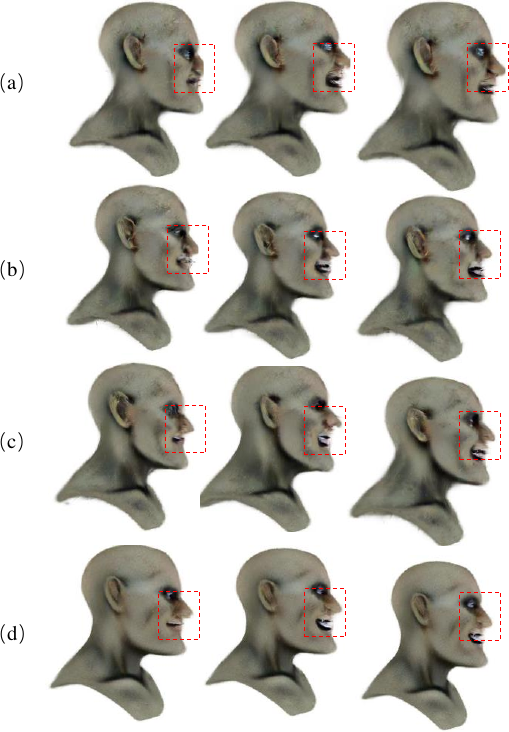}         
    \caption{\textbf{Ablation study validating the feature bank.} (a) No feature bank is used. (b) Only the key-value feature bank is utilized. (c) Only the output feature bank is used. (d) Our full implementation with both key-value and output feature banks. (see supplementary videos for better comparisons)}
    \label{fig:fb}     
    \vspace{-3mm}   
\end{figure}

\begin{figure}
    \centering
    \includegraphics[width=1\linewidth]{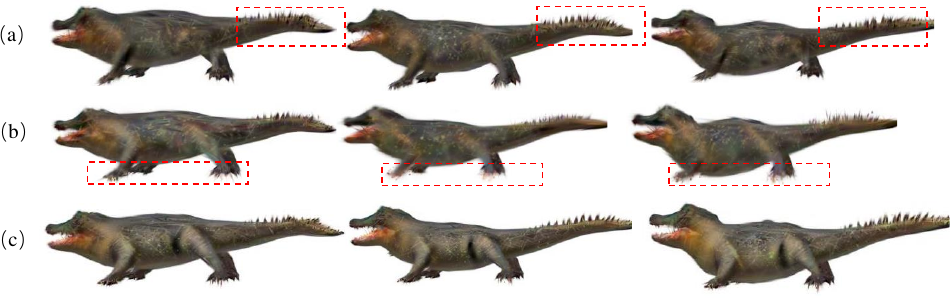}
    \caption{\textbf{Ablation study on the feature bank updating method.} (a) Updating the feature bank using a queue with a length of 1. (b) Updating the feature bank using a queue with a length of 2. (c) Dynamically updating the feature bank (our approach).}
    \label{fig:queue}
\end{figure}

\begin{figure}
    \centering
    \includegraphics[width=0.6\linewidth]{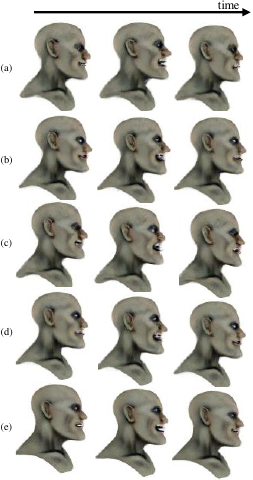}
    \caption{\textbf{Ablation study on using the feature bank at different network blocks.} (a) Without using the feature bank. (b) Using the feature bank at downsampling blocks. (c) Using the feature bank at middle blocks. (d) Using the feature bank at upper blocks. (e) Using the feature bank at all blocks.}
    \label{fig:net}
\end{figure}

\begin{figure}
    \centering
    \resizebox{\linewidth}{!}{\includegraphics{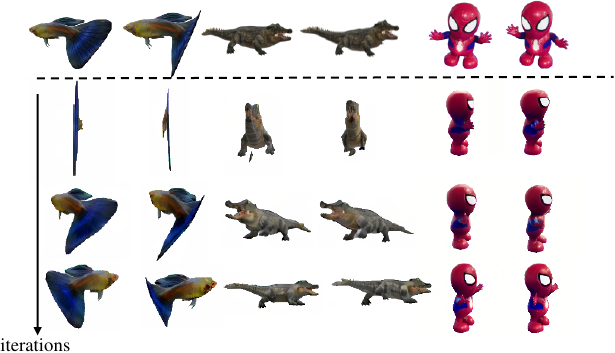}}
    \caption{\textbf{Ablation study on the number of iterations}}
    \label{fig:iterb}
\end{figure}

\begin{figure}
    \centering
    \includegraphics[width=0.7\linewidth]{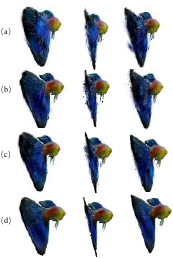}
    \caption{\textbf{Ablation study on the progressively iterations and our usage of feature bank during multiple iterations.} (a) Randomly selecting viewpoints without feature bank interaction across multiple iterations.(b) Randomly selecting viewpoints with feature bank interaction across multiple iterations.(c) Progressively iterating without feature bank interaction across multiple iterations.(d) Our proposed approach, which incorporates both.}
    \label{fig:itera}
\end{figure}
\textbf{Ablation study on the utilization of our feature bank} To better visualize the role of the key-value feature bank and the output feature bank, we conducted a comprehensive controlled experiment, as shown in Fig.~\ref{fig:fb}. Our full implementation, which incorporates both the key-value and output feature banks, achieves the best consistency.

\textbf{Ablation study on the feature bank updating method.} We conducted a detailed comparison of different feature bank updating methods. As shown in Figure \ref{fig:queue} (a), using a queue with a window size of 1 for updates may result in insufficient utilization of past information, leading to inconsistencies in local details over time. In contrast, Figure \ref{fig:queue} (b) demonstrates that simply increasing the window size can lead to suboptimal fusion of information, where excessive and redundant data introduce confusion, ultimately degrading the quality of the output. Our proposed method, illustrated in Figure \ref{fig:queue} (c), effectively integrates past information, enhancing the amount of useful information and thereby improving the overall quality of the generated results.

\textbf{Ablation study on using the feature bank at different network blocks. } 
We examine the impact of incorporating a feature bank at different network blocks. As shown in Figure~\ref{fig:net}, using feature bank across all blocks (e) yields the best performance with minimal artifacts, demonstrating its effectiveness in preserving essential features. Applying it to specific blocks—downsampling (b), middle (c), or upper (d)—still provides benefits but to a lesser extent. Without feature bank (a), the model struggles to retain rich information, leading to weaker representations and more artifacts.

\textbf{Ablation study on the progressively iterations and our usage of the feature bank during multiple iterations.} We analyze the impact of progressive iterations and feature bank interactions using four configurations. As shown in Figure~\ref{fig:itera}, randomly selecting viewpoints without feature bank interaction (a) leads to the poorest performance due to a lack of accumulated information. Feature bank interaction alone (b) improves quality but lacks progressive refinement. Progressive iterations (c) enhance stability but underutilize historical information. Our method (d), combining both, achieves the best performance by balancing temporal consistency and information integration.

\textbf{Study on the number of iterations.} To better illustrate the benefits of multiple iterations, we visualize the multi-view sequences outputted by stage 1 in the pipeline. As shown in the figure \ref{fig:iterb}, during the first iteration, information from side and other views is noticeably incomplete. However, as the number of iterations increases, we progressively generate additional views that are similar to the missing viewpoints, effectively compensating for the lack of information. Furthermore, with each iteration, the generated images maintain improved spatial consistency compared to previous iterations.

\section{Computational cost}

To optimize memory usage, we frequently perform tensor operations between the CPU and GPU, allowing the generation of multi-view image sequences with a memory footprint of approximately 14GB. This approach effectively maximizes the retention of historical information while minimizing additional memory overhead, compared to the initial implementation's 10GB \cite{zeng2025stag4d}. However, it comes with a notable time penalty: inference over 75 steps takes about 60 seconds, whereas the baseline (STAG4D) achieves this in just 10 seconds. Moreover, calculating the clip score similarity adds an additional 15 minutes of processing time per iteration. For a sequence of 32 images and three iterations, the total generation time reaches approximately 150 minutes. While the inclusion of a feature bank boosts performance, it also increases both memory and computational demands.


\end{document}